\documentclass[10pt,twocolumn,letterpaper]{article}

\usepackage{iccv}
\usepackage{times}
\usepackage{epsfig}
\usepackage{graphicx}
\usepackage{subcaption}
\usepackage{comment}
\usepackage{amsmath}
\usepackage{amssymb}
\usepackage{booktabs}
\usepackage{multicol}

\usepackage{multirow}
\usepackage{algorithm, algorithmic}

\DeclareMathOperator*{\mlp}{MLP}
\DeclareMathOperator*{\fc}{FC}

\newcommand{\co}{n} 
\newcommand{\bbox}{\mathbf{b}} 
\newcommand{\bboxes}{\mathbb{B}} 
\newcommand{\lset}{\mathbb{L}} 
\newcommand{\lsetvec}{\mathbf{s}} 
\newcommand{\lab}{\mathbf{l}} 

\usepackage{pifont}
\newcommand{\cmark}{\ding{51}}%

\usepackage{xr}

\usepackage[pagebackref=true,breaklinks=true,letterpaper=true,colorlinks,bookmarks=false]{hyperref}
\usepackage{pdfcomment}

\iccvfinalcopy 

\def\iccvPaperID{1245} 

\begin{document}

\title{LayoutVAE: Stochastic Scene Layout Generation From a Label Set}

\author{Akash Abdu Jyothi$^{1,3}$, Thibaut Durand$^{1,3}$, Jiawei He$^{1,3}$,  Leonid Sigal$^{2,3}$, Greg Mori$^{1,3}$ \\
$^{1}$Simon Fraser University \qquad $^2$University of British Columbia \qquad $^{3}$Borealis AI\\
{\tt\small \{aabdujyo, tdurand, jha203\}@sfu.ca}\qquad
{\tt\small{lsigal}@cs.ubc.ca}\qquad
{\tt\small{mori}@cs.sfu.ca}
}

\maketitle

\begin{abstract}
Recently there is an increasing interest in scene generation within the research community. However, models used for generating scene layouts from textual description largely ignore plausible visual variations within the structure dictated by the text. We propose LayoutVAE, a variational autoencoder based framework for generating stochastic scene layouts. LayoutVAE is a versatile modeling framework that allows for generating full image layouts given a label set, or per label layouts for an existing image given a new label. In addition, it is also capable of detecting unusual layouts, potentially providing a way to evaluate layout generation problem. Extensive experiments on MNIST-Layouts and challenging COCO 2017 Panoptic dataset verifies the effectiveness of our proposed framework.
\end{abstract}

\section{Introduction}
\label{sec:introduction}

Scene generation, which usually consists of realistic generation of multiple objects under a semantic layout, remains one of the core frontiers of computer vision. Despite the rapid progress and recent successes in object generation (\eg, celebrity face, animals, \etc) \cite{bodla2018,he2018,karras2018progressive} and scene generation \cite{chen2017,hong2018,johnson2018,li2019object,park2019,zhang2017,zhao2018}, little attention has been paid to frameworks designed for stochastic semantic layout generation.
Having a robust model for layout generation will not only allow us to generate reliable scene layouts, but also provide priors and means to infer latent relationships between objects, advancing progress in the scene understanding domain.

A plausible semantic layout calls for reasonable \textit{spatial} and \textit{count} relationships (relationships between the number of instances of different labels) between objects in a scene \cite{choi2010exploiting,sudderth2005learning}. For example, a person would either ride (\textit{on top of}) a horse, or stand \textit{next to} a horse, but seldom would he be \textit{under} a horse. Another example would be that the number of ties would very likely be smaller than or equal to the number of people in an image. The first example shows an instance of a plausible spatial relationship and the second shows a case of a generic count relationship. 
Such intrinsic relationships buried in high-dimensional visual data are usually learned implicitly by mapping the textual description to visual data. 
However, since the textual description can always be treated as an abstraction of the visual data, the process becomes a \textit{one-to-many} mapping. In other words, given the same text information as a condition, a good model should be able to generate multiple plausible images all of which satisfy the semantic description. 

\begin{figure}[t]
\centering
\begin{tabular}{cc}
\includegraphics[height=2.6cm, width=4cm]{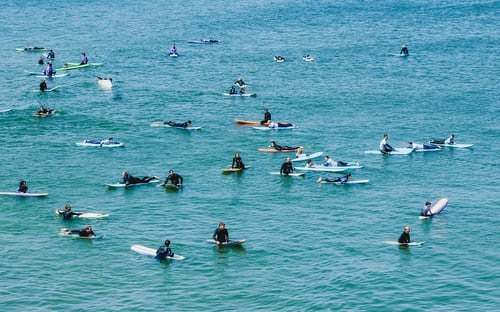}
\includegraphics[height=2.6cm, width=4cm]{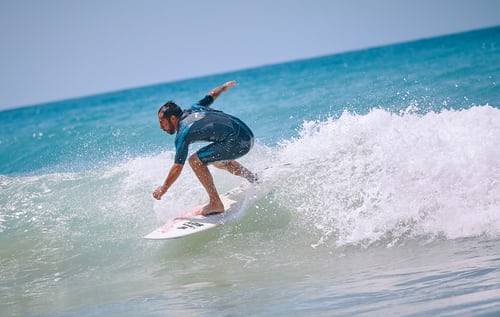}
\\
\includegraphics[height=2.6cm, width=4cm]{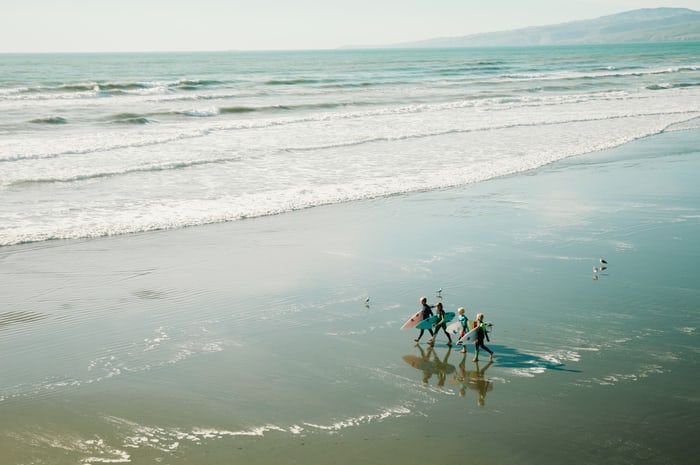}
\includegraphics[height=2.6cm, width=4cm]{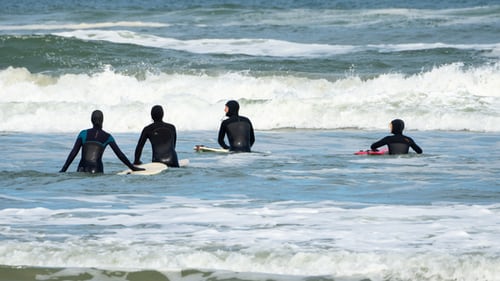}
\end{tabular}
\caption{Several images with the same label set: \textit{person}, \textit{surfboard} and \textit{sea}. Given this simple label set, we observe that a large and diverse set of layouts is plausible.}
\label{fig:fig_1}
\end{figure}

Previous works focused on a popular simplified instance of the problem described above: scene generation based on sentence description \cite{deng2018,hong2018,johnson2018,li2019object,reed2016learning,reed2016generative,zhang2017}. A typical sentence description includes partial information on both the background and objects, along with details of the objects' appearances and scene layout. These frameworks rely heavily on the extra relational information provided by the sentence. As a result, although these methods manage to generate realistic scenes, they tend to ignore learning the intrinsic relationships between the objects, prohibiting the wide adoption of such models where weaker descriptions are provided.

In this work, we consider a more sophisticated problem: scene generation based on a label set description. 
A label set, as a much weaker description, only provides the set of labels present in the image (without any additional relationship description), requiring the model to learn \textit{spatial} and \textit{count} relationships from visual data.

Furthermore, the ambiguity of this type of label set supervision calls for diverse scene generation. 
For example, given the label set \textit{person, surfboard, sea}, a corresponding scene could have multiple instances of each label (under plausible count relationships), positioned at various locations (under plausible spatial relationships). For instance in the COCO dataset \cite{lin2014microsoft}, there are 869 images in the training set that have the label set \textit{person}, \textit{sea} and \textit{surfboard}.
\autoref{fig:fig_1} shows examples of multiple plausible images with this label set.

We propose LayoutVAE, a stochastic model capable of generating scene layouts given a label set. 
The proposed framework can be easily embedded into existing scene generation models that take scene layout as input, such as \cite{hinz2019generating,zhao2018}, providing them plausible and diverse layouts. Our main contributions are as follows.

\begin{itemize}
    \item We propose a new model for stochastic scene layout generation given a label set input. Our model has two components, one to model the distributions of count relationships between objects and another to model the distributions of spatial relationships between objects.
    \item We propose a new synthetic dataset, MNIST-Layouts, that captures the stochastic nature of scene layout generation problem.
    \item We experimentally validate our model using MNIST-Layouts and the COCO \cite{lin2014microsoft} dataset which contains complex real world scene layouts. We analyze our model and show that it can be used to detect unlikely scene layouts.
\end{itemize}

\section{Related Work}
\label{sec:related}

\noindent \textbf{Sentence-conditioned image generation.}
A variety of models have proposed to generate an image given a sentence.
Reed~\etal~\cite{reed2016generative} use a GAN~\cite{goodfellow2014} that is conditioned on a text encoding for generating images. Zhang~\etal~\cite{zhang2017} propose a GAN based image generation framework where the image is progressively generated in two stages at increasing resolutions. Reed~\etal~\cite{reed2016learning} perform image generation with sentence input along with additional information in the form of keypoints or bounding boxes. 

Hong~\etal~\cite{hong2018} break down the process of generating an image from a sentence into multiple stages. The input sentence is first used to predict the objects that are present in the scene, followed by prediction of bounding boxes, then semantic segmentation masks, and finally the image. While scene layout generation in this work predicts probability distributions for bounding box layout, it fails to model the stochasticity intrinsic in predicting each bounding box.
Gupta~\etal~\cite{gupta2018} use an approach similar to \cite{hong2018} to predict layouts for generating videos from scripts.
Johnson~\etal~\cite{johnson2018} uses the scene graph generated from the input sentence as input to the image generation model. 
Given a scene graph, their model can generate only one scene layout.

Deng~\etal~\cite{deng2018} propose PNP-Net, a VAE framework to generate image of an abstract scene from a text based program that fully describes it. 
While PNP-Net is a stochastic model for generation, it was tested on synthetic datasets with only a small number of classes. Furthermore, it tries to encode the entire image into a single latent code whereas, in LayoutVAE, we break down just the layout generation step into two stages with multiple steps in each stage. Based on these reasons, it is unclear whether PNP-Net can scale up to real world image datasets with a large number of classes.
Tao~\etal~\cite{tao18} propose a GAN based model with attention for sentence to image generation. 
The more recent work from Li~\etal~\cite{li2019object} follow a multi-stage approach similar to \cite{hong2018} to generate an image from a sentence, with the key novelty of using attention mechanisms to create more realistic objects in the image.

\noindent \textbf{Layout generation in other contexts.} Chang~\etal~\cite{chang2014spatial} propose a method for 3D indoor scene generation based on text description by placing objects from a 3D object library, and is later improved in ~\cite{chang2015lexground} by learning the grounding of more detailed text descriptions with 3D objects.
Wang~\etal~\cite{wang2018deep} use a convolutional network that iteratively generates a 3D room scene by adding one object at a time.
Qi~\etal~\cite{qi2018human} propose a spatial And-Or graph to represent indoor scenes, from which new scenes can be sampled. Different from most other works, they use human affordances and activity information with respect to objects in the scene to model probable spatial layouts. 
Li~\etal~\cite{li2018grains} propose a VAE based framework that encodes object and layout information of indoor 3D scenes in a latent code. During generation, the latent code is recursively decoded to obtain details of individual objects and their layout.

More recently, Li~\etal~\cite{li2018layoutgan} proposed LayoutGAN, a GAN based model that generates layouts of graphic elements (rectangles, triangles etc.). 
While this work is close to ours in terms of the problem focus, LayoutGAN generates label sets based on input noise, and it cannot generate layout for a given set of labels.

\noindent \textbf{Placing objects in scenes.} Lee~\etal~\cite{lee2018} propose a conditional GAN model for the problem of adding the segmentation mask of a new object to the semantic segmentation of an image. Lin~\etal~\cite{lin2018} address a similar problem of adding an RGB mask of an object into a background image.

\section{Background}

\begin{figure*}[t]
\centering
\includegraphics[width=\textwidth]{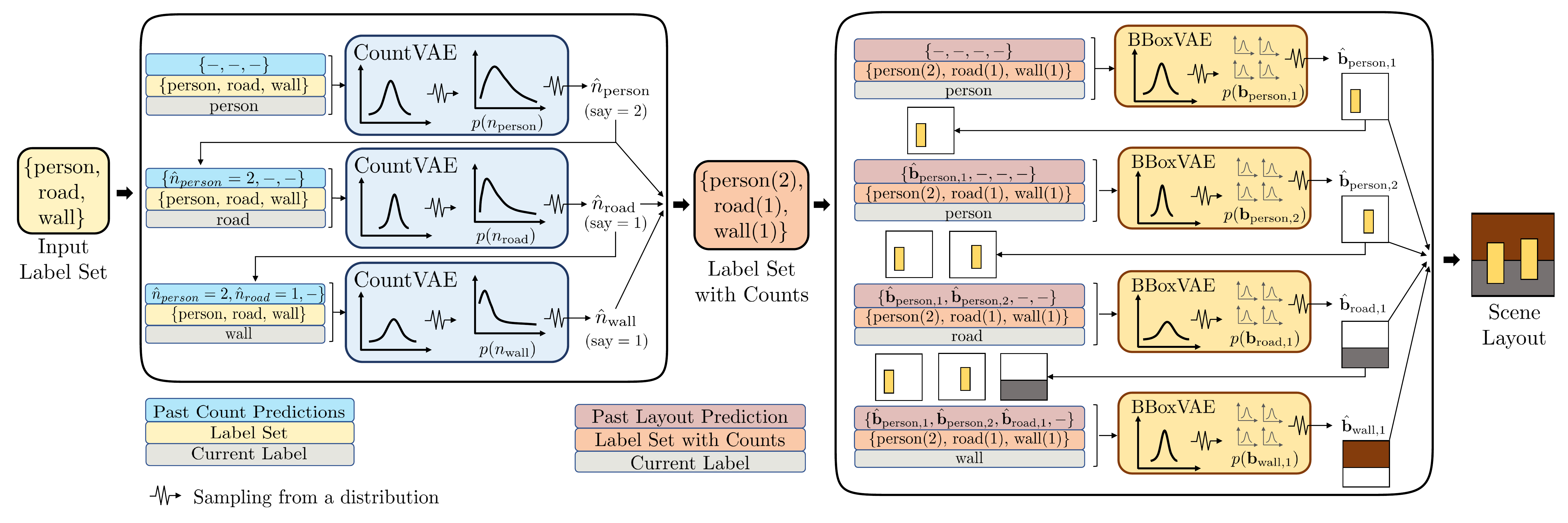}
\caption{\textbf{Model}. LayoutVAE is composed of two models: CountVAE which predicts the number of objects for each category and BBoxVAE which predicts the bounding box of each object. A graphical model is given in \autoref{app:graphical_model} of the supplementary.}
\label{fig:model_structure}
\end{figure*}

In this section, we first define the problem of scene layout generation from a label set, and then provide an overview of the base models that LayoutVAE is built upon.


\subsection{Problem Setup} 
We are interested in modeling contextual relationships between objects in scenes,
and furthermore generation of diverse yet plausible scene layouts given a label set as input. The problem can be formulated as follows.

Given a collection of $M$ object categories, we represent the label set corresponding to an image in the dataset as $\lset \subseteq \{1,2,3,...,M\}$ which indicates the categories present in the image.
Note that here we use the word ``object" in its very general form: \textit{car}, \textit{cat}, \textit{person}, \textit{sky} and \textit{water} are objects.
For each label $k \in \lset$, let $\co_k$ be the number of objects of that label in the image and $\bboxes_k=\{\bbox_{k,1}, \bbox_{k,2}, ... , \bbox_{k,\co_k}\}$ be the set of bounding boxes.
$\bbox_{k,i}=[x_{k,i}, y_{k,i}, w_{k,i}, h_{k,i}]$ represents the top-left coordinates, width and height of the $i$-th bounding box of category $k$.
We train a generative model to predict diverse yet plausible sets of $\{\bboxes_k : k \in  \lset\}$ given the label set $\lset$ as input.

\subsection{Base Models}
\noindent \textbf{Variational Autoencoders.} 
A variational autoencoder (VAE)~\cite{Kingma2014} describes an instance of a family of generative models $p_{\theta}(\mathbf{x},\mathbf{z}) = p_{\theta}(\mathbf{x}|\mathbf{z})p_\theta(\mathbf{z})$ with a complex likelihood function $p_{\theta}(\mathbf{x}|\mathbf{z})$ and an amortized inference network $q_{\phi}(\mathbf{z}|\mathbf{x})$ to approximate the true posterior $p_{\theta}(\mathbf{z}|\mathbf{x})$. Here $\mathbf{x}$ represents observable data examples, $\mathbf{z}$ the latent codes, $\theta$ the generative model parameters, and $\phi$ the inference network parameters.
To prevent the latent variable $\mathbf{z}$ from just copying $\mathbf{x}$, we force $q_{\phi}(\mathbf{z}|\mathbf{x})$ to be close to the prior distribution $p_\theta(\mathbf{z})$ using a KL-divergence term.
Usually in VAE models, $p_\theta(\mathbf{z})$ is a fixed Gaussian $\mathcal{N}(0,I)$.
Both the generative and the inference networks are realized as non-linear neural networks. 
An evidence lower bound (ELBO) $\mathcal{L}(\mathbf{x};\theta,\phi)$ on the generative data likelihood $\log p(\mathbf{x})$ is used to jointly optimize $\theta$ and $\phi$:
\begin{equation}
\mathcal{L}(\mathbf{x};\theta,\phi) = \mathbb{E}_{q_{\phi}(\mathbf{z}|\mathbf{x})} \left[\log p_{\theta}(\mathbf{x}|\mathbf{z})\right] - \textrm{KL} \left(q_{\phi}(\mathbf{z}|\mathbf{x})||p_\theta(\mathbf{z})\right)
\label{eq:ELBO}
\end{equation}

\noindent \textbf{Conditional VAEs.} A conditional VAE (CVAE)~\cite{sohn2015} defines an extension of VAE that conditions on an auxiliary description $\mathbf{c}$ of the data. The auxiliary conditional variable makes it possible to infer a conditional posterior $q_\phi(\mathbf{z}|\mathbf{x},\mathbf{c})$ as well as perform generation $p_\theta(\mathbf{x}|\mathbf{z},\mathbf{c})$ based on a given description $\mathbf{c}$. The ELBO is thus updated as:
\begin{equation}
\begin{split}
\mathcal{L}^{CVAE}(\mathbf{x},\mathbf{c};\theta,\phi) = & \mathbb{E}_{q_{\phi}(\mathbf{z}|\mathbf{x},\mathbf{c})} \left[\log p_{\theta}(\mathbf{x}|\mathbf{z},\mathbf{c})\right] \\
& - \textrm{KL} \left(q_{\phi}(\mathbf{z}|\mathbf{x},\mathbf{c})||p_\theta(\mathbf{z}|\mathbf{c})\right)
\end{split}
\label{eq:ELBO}
\end{equation}
In CVAE models, the prior of the latent variables $\mathbf{z}$ is modulated by the auxiliary input $\mathbf{c}$. 

\section{LayoutVAE for Stochastic Scene Layout Generation}
In this section, we present LayoutVAE and describe the scene layout generation process given a label set. 
As discussed in \autoref{sec:introduction}, this task is challenging and solving it requires answering the following two questions: \textit{what is the number of objects for each category?} and \textit{what is the location and size for each object?}
LayoutVAE is naturally decomposed into two models: one to predict the count for each given label, named \textit{CountVAE}, and another to predict the location and size of each object, named \textit{BBoxVAE}. 
The overall structure of the proposed LayoutVAE is shown in \autoref{fig:model_structure}. 
The number of objects (count) for each label is first predicted by CountVAE, then BBoxVAE predicts the bounding box for each object. 
The two-step approach with stochastic models naturally allows LayoutVAE to generate diverse layouts. In addition, it provides the flexibility to handle various types of input as it allows us to use each module independently. For example, BBoxVAE can be used to generate a layout if counts are available, or add a single bounding box in an existing image given a new label. 

The input to CountVAE is the set of labels $\lset$ and it predicts the distributions of object counts $\{\co_k : k \in \lset\}$ autoregressively, where $\co_k$ is the object count for category $k$. 
The input of BBoxVAE is the set of labels along with the counts for each of the label ${\{\co_k : k\in \lset\}}$ and it predicts the distribution of each bounding box $\bbox_{k,i}$ autoregressively.



\subsection{CountVAE}

CountVAE is an instance of conditional VAE designed to predict conditional count distribution for the labels in an autoregressive manner. We use a predefined order for the label set (we observe empirically that a predefined order is superior to randomized order across samples; learning an order is a potential extension but adds complexity). 
In practice, CountVAE predicts the count of the first label, then the count of the second label \etc, at each step conditioned on already predicted counts.
It models the distribution of count $\co_k$ given the label set $\lset$, the current label $k$ and the counts for each category that was predicted before $\{\co_m : m < k\}$.
The conditioning input for CountVAE is: 
\begin{align}
\mathbf{c}^c_k= \left(\lset, k, \{\co_m : m < k\} \right)
\label{eq:condinput_countvae}
\end{align}
where $(\cdot,\cdot)$ denotes a tuple.
We use the notation of superscript $c$ to indicate that it is related to the CountVAE. 
We use a Poisson distribution to model the number of occurrences of the current label at each step:
\begin{equation}
p_{\theta^c}(\co_k | \mathbf{z}^c_k, \mathbf{c}^c_k) = \frac{(\lambda(\mathbf{z}^c_k, \mathbf{c}^c_k))^{(\co_k-1)} e^{-\lambda(\mathbf{z}^c_k, \mathbf{c}^c_k)}}{(\co_k-1)!}
\label{eq:poisson}
\end{equation}
where $p_{\theta^c}(\co_k | \mathbf{z}^c_k, \mathbf{c}^c_k)$ is a probability distribution over random variable $\co_k$, $\theta^c$ represents the generative model parameters of CountVAE, and $\lambda(\mathbf{z}^c_k, \mathbf{c}^c_k)$ is the rate which depends on the latent variable sample  $\mathbf{z}^c_k$ and the conditional input $\mathbf{c}^c_k$. 
Note that we learn the distribution over $\co_k-1$ since the count per label is always 1 or greater in this problem setting.

The latent variable is sampled from the approximate posterior during learning and from the conditional prior during generation.
The latent variables model the ambiguity in the scene layout.
Both approximate posterior and prior are modeled as multivariate Gaussian distributions with diagonal covariance with parameters as shown below: 
\begin{align}
q_{\phi^c}(\mathbf{z}_k^c|\co_k,\mathbf{c}_k^c) &= \mathcal{N}(\mu_{\phi^c}(\co_k,\mathbf{c}_k^c), \sigma^2_{\phi^c}(\co_k,\mathbf{c}_k^c))
\\
p_{\theta^c}(\mathbf{z}_k^c|\mathbf{c}_k^c) &= \mathcal{N}(\mu_{\theta^c}(\mathbf{c}_k^c),\sigma^2_{\theta^c}(\mathbf{c}_k^c))
\end{align}
where $\mu_{\phi^c}$ (resp. $\mu_{\theta^c}$) and $\sigma^2_{\phi^c}$ (resp. $\sigma^2_{\theta^c}$) are functions that estimate the mean and the variance of the approximate posterior (resp. prior). Details of the model architecture are given in \autoref{app:countVAE} of the supplementary.

\begin{algorithm}[t]
\caption{CountVAE: Loss computation on the image}
\label{alg:countvae_learning}
\begin{algorithmic}[1]
\REQUIRE Label set $\lset$, instance count ${\{\co_k : k \in \lset\}}$ 
\STATE $\mathcal{L}^{Count} = 0$
\FOR{$k \in \lset$} 
\STATE Compute the conditioning input $\mathbf{c}^c_k$  for category $k$ (\autoref{eq:condinput_countvae})   
\STATE Compute the variational lower bound $\mathcal{L}^c(\co_k,\mathbf{c}_k^c;\theta^c,\phi^c)$ (\autoref{eq:elbo_countvae})
\STATE $\mathcal{L}^{Count} = \mathcal{L}^{Count} + \mathcal{L}^c(\co_k,\mathbf{c}_k^c;\theta^c,\phi^c)$
\ENDFOR
\ENSURE $\mathcal{L}^{Count} \big/ |\lset|$
\end{algorithmic}
\end{algorithm}

\paragraph{Learning.} 
The model is optimized by maximizing the ELBO over ${\{\co_k : k \in \lset\}}$. 
The ELBO corresponding to the label count $\co_k$ is given by 
\begin{align}
\!\!\mathcal{L}^c(\co_k,\mathbf{c}_k^c; \theta^c \!\!, \phi^c) = & \mathbb{E}_{q_{\phi^c}(\mathbf{z}_k^c|\co_k,\mathbf{c}_k^c)} \left[\log p_{\theta^c}(\co_k|\mathbf{z}_k^c,\mathbf{c}_k^c)\right] \label{eq:elbo_countvae} \\
& - \textrm{KL} \left(q_{\phi^c}(\mathbf{z}_k^c|\co_k,\mathbf{c}_k^c)||p_{\theta^c}(\mathbf{z}_k^c|\mathbf{c}_k^c)\right) 
\nonumber
\end{align}
where $\phi^c$ represents the inference model parameters of CountVAE. 
Log-likelihood of the ground truth count under the predicted Poisson distribution is used to compute $p_{\theta^c}(\co_k|\mathbf{z}_k^c,\mathbf{c}_k^c)$, while the KL divergence between the two Gaussian distributions is computed analytically. 
The computation of the loss for the label set $\lset$ is given in \autoref{alg:countvae_learning}.

\paragraph{Generation.}
Given a label set $\lset$, the CountVAE autoregressively predicts the object count for each category by sampling from the count distribution (\autoref{eq:poisson}).
We now present the generation process to predict the object count of category $k$. 
We first compute the conditional input $\bar{\mathbf{c}}^c_k$:
\begin{align}
\bar{\mathbf{c}}^c_k = \left( \lset, k, \{\hat{\co}_m: m < k\} \right)
\end{align}
where $\hat{\co}_m$ is the predicted instance count for category $m$.
Note that to predict the instance count of category $k$, the model exploits the predicted counts $\hat{\co}_m$ for previous categories $m < k$ to get more consistent counts. 
Then, we sample a latent variable $\hat{\mathbf{z}}_k^c$ from the conditional prior:
\begin{align}
\hat{\mathbf{z}}_k^c \sim p_{\theta^c}(\mathbf{z}_k^c|\bar{\mathbf{c}}_k^c) 
\end{align}
Finally, the count is sampled from the predicted Poisson count distribution:
\begin{align}
\hat{\co}_k \sim  p_{\theta^c}(\co_k|\hat{\mathbf{z}}_k^c,\bar{\mathbf{c}}_k^c) 
\end{align}
This label count is further used in the conditioning variable for the future steps of CountVAE. 

\subsection{BBoxVAE}

Given the label set $\lset$ and the object count per category $\{\co_m : m \in \lset\}$, the BBoxVAE predicts the distribution of coordinates for the bounding boxes autoregressively. We follow the same predefined label order as CountVAE in the label space, and order the bounding boxes from left to right for each label. All bounding boxes for a given label are predicted before moving on to the next label. 
BBoxVAE is a conditional VAE that models the $j^\text{th}$ bounding box $\bbox_{k,j}$ for the label $k$ given the label set $\lset$ along with the count for each label, current label $k$ and coordinate and label information of all the bounding boxes that were predicted earlier. Previous predictions include all the bounding boxes for previous labels as well as the ones previously predicted for the current label:
$\bboxes_{k,j}^{prev} = \{\bbox_{m,:}: m < k\} \cup \{\bbox_{k,i}: i < j\}$
The conditioning input of the BBoxVAE is: 
\begin{align}
\mathbf{c}^b_{k,j} = \Big(\{\co_m : m \in \lset\}, k,
\bboxes_{k,j}^{prev} \Big) 
\label{eq:condinput_bboxvae}
\end{align}
We use the notation of exponent $b$ to indicate that it is related to the BBoxVAE. 
We model bounding box coordinates and size information using a quadrivariate Gaussian distribution as shown in \autoref{eq:gaussian}. We omit the subscript $k,j$ for all variables in the equation for brevity.
\begin{equation}
p_{\theta^b}(x,y,w,h | \mathbf{z}^b\!, \mathbf{c}^b) \!=\! \mathcal{N}\big(x,y,w,h | \mu^b(\mathbf{z}^b\!, \mathbf{c}^b), \sigma^b(\mathbf{z}^b\!, \mathbf{c}^b)\big)
\label{eq:gaussian}
\end{equation}
where $\theta^b$ represents the generative model parameters of BBoxVAE, ${\mathbf{z}^b}$ denotes a sampled latent variable and $\mu^b$ and $\sigma^b$ are functions that estimate the mean and the variance respectively for the Gaussian distribution. 
$x$ (resp. $y$) is the $x$- (resp. $y$-) coordinate of the top left corner and $w$ (resp. $h$) is the width (resp. height) of the bounding box.
These variables are normalized between 0 and 1 to be independent of the dimensions of the image.
Details of the model architecture are given in \autoref{app:BBoxVAE} of the supplementary.

\begin{algorithm}[t]
\caption{BBoxVAE: Loss computation on the image}
\label{alg:bboxvae_learning}
\begin{algorithmic}[1]
\REQUIRE Label set $\lset$, instance count ${\{\co_k : k \in \lset\}}$, set of bounding boxes $\{\bboxes_k : k \in \lset\}$
\STATE $\mathcal{L}^{BBox} = 0$
\FOR{$k \in \lset$} 
\FOR{$j \in \{1, \ldots, \co_k\}$} 
\STATE Compute the conditioning input $\mathbf{c}^b_{k,j}$  for the $j$-th bounding box of category $k$ (\autoref{eq:condinput_bboxvae})
\STATE Compute the variational lower bound $\mathcal{L}^b(\bbox_{k,j},\mathbf{c}^b_{k,j};\theta^b,\phi^b)$ (\autoref{eq:elbo_bboxvae})
\STATE $\mathcal{L}^{BBox} = \mathcal{L}^{BBox} + \mathcal{L}^b(\bbox_{k,j},\mathbf{c}^b_{k,j};\theta^b,\phi^b)$
\ENDFOR
\ENDFOR
\ENSURE $\mathcal{L}^{BBox}  \big/ \big(\textstyle \sum_{k \in \lset} n_k\big)$
\end{algorithmic}
\end{algorithm}

\paragraph{Learning.}
We train the model in an analogous manner to CountVAE by maximizing the variational lower bound over the entire set of bounding boxes. For the bounding box $\bbox_{k,j}$, the lower bound is given by (omitting the subscript $k,j$ for all variables in the equation):
\begin{align}
\mathcal{L}^b(\bbox,\mathbf{c}^b;\theta^b,\phi^b) = & \mathbb{E}_{q_{\phi^b}(\mathbf{z}^b|\bbox,\mathbf{c}^b)} \left[\log p_{\theta^b}(\bbox|\mathbf{z}^b,\mathbf{c}^b)\right] 
\label{eq:elbo_bboxvae} \\
& - \textrm{KL} \left(q_{\phi^b}(\mathbf{z}^b|\bbox,\mathbf{c}^b)||p_{\theta^b}(\mathbf{z}^b|\mathbf{c}^b)\right)
\nonumber
\end{align}
where $\phi^b$ represents the inference model parameters of BBoxVAE. 
The computation of the loss for all the bounding boxes is given in \autoref{alg:bboxvae_learning}.


\paragraph{Generation.}
Generation for BBoxVAE is performed in an analogous manner as CountVAE. 
Given a label set along with the number of instances of each label, BBoxVAE autoregressively predicts bounding box distributions (\autoref{eq:gaussian}) by sampling a latent variable from the conditional prior $p_{\theta^b}(\mathbf{z}^b|\mathbf{c}^b)$.
The conditioning variable is updated after each step by including a sampled bounding box from the current step prediction.

\section{Experiments}

We implemented LayoutVAE in PyTorch, and used Adam optimizer \cite{Kingma2015} for training the models.
CountVAE was trained for $50$ epochs at a learning rate of $10^{-5}$ with batch size $32$. 
BBoxVAE was trained for $150$ epochs at a learning rate of $10^{-4}$ with batch size $32$. We chose the best model during training based on the validation loss, and evaluate that model's performance on the test set.
We present quantitative and qualitative results that showcase the usefulness of LayoutVAE model. 

\subsection{Datasets}

\begin{figure}
\centering
\includegraphics[width=0.45\textwidth]{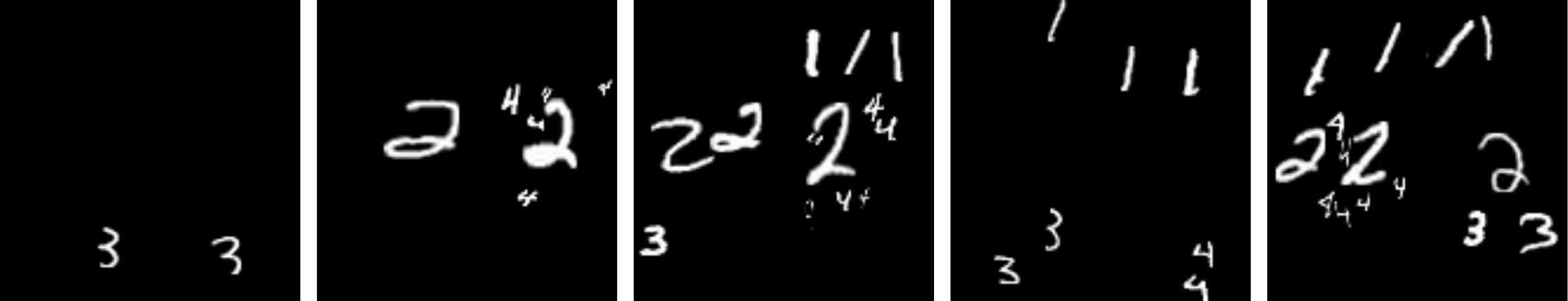}
\vspace{-3mm}
\caption{\textbf{Samples from MNIST-Layouts dataset.} We randomly sample MNIST digits of appropriate labels to fill in the bounding boxes. The rules for count and spatial layout (e.g.\ large 2s in the middle, small 3s at the bottom etc.\ ) are described in \autoref{sec:supp_mnist_layout} in the supplementary.}
\vspace{-3mm}
\label{fig:mnist_train}
\end{figure}

\vspace{0.05in}
\noindent \textbf{MNIST-Layouts.}
We created this dataset by placing multiple MNIST digits on a $128\times128$ canvas based on predefined rules detailed in \autoref{sec:supp_mnist_layout} in the supplementary material. 
We use the digits $\{1,2,3,4\}$ as the global set of labels, thus limiting the maximum number of labels per image to $4$.
The dataset consists of $5000$ training images, and $1000$ images each for validation and testing. 
\autoref{fig:mnist_train} shows some examples of scene layouts from the training set. 

\vspace{0.05in}
\noindent \textbf{COCO.}
We use 2017 Panoptic version of COCO dataset \cite{lin2014microsoft} for our experiments. This dataset has images from complex everyday scenes with bounding box annotations for all major \textit{things} and \textit{stuffs} present in the scene. 
The official release has $118,267$ annotated images in the training set and $5000$ in the validation set. 
To allow a comparison with future methods, we use the official validation set as test set, and use the last $5000$ images from the training set as validation set in our experiments. 
This dataset has $80$ \textit{thing} categories (person, table, chair \etc) and $53$ \textit{stuff} categories (sky, sand, snow \etc). 
The largest number of labels present in a label set is $34$, and the largest number of bounding box instances present in an image is $98$. 
We normalize all bounding box dimensions to $[0,1]$ by dividing by the image size (width or height, as appropriate). This allows the model to predict layouts for square images of any size.
We ignore \textit{thing} instances that are tagged \textit{``iscrowd"} \ie consists of a group of thing instances. They account for less than $1$ percent of all the annotations.

\begin{table}[t]
\centering
\begin{tabular}{lcccc}
\toprule
\multirow{3}{*}{Model} & \multicolumn{2}{c}{Label Count} &  BBox \\
\cmidrule(lr){2-4}
 & \multirow{2}{*}{\shortstack[c]{Accuracy \\ (\%)}} & \multirow{2}{*}{\shortstack[c]{Accuracy \\ within $\pm1$ (\%)}} &  \multirow{2}{*}{\shortstack[c]{Mean \\ IoU}} \\
\\
\midrule
AR-MLP & $74.21$ & $87.50$ & $0.18$   \\
BBoxLSTM & - & - & $0.17$  \\
BLSTM & $73.98$ & $87.87$ & $0.15$ \\
sg2im~\cite{johnson2018} & - & - & $0.14$ \\ 
LayoutVAE & $\mathbf{78.38}$ & $\mathbf{89.87}$ & $\mathbf{0.20}$ \\
\bottomrule
\end{tabular}
\vspace{-3mm}
\caption{\textbf{Comparison with baselines on COCO dataset using accuracy metrics.} We use the most likely count or bounding box 
as the prediction for all the models. While all models predict a distribution as output, only LayoutVAE has a stochastic latent code. Mean of the latent code distribution is used to obtain the output (count or bounding box) distribution for measuring the accuracy of LayoutVAE.
}
\label{tab:test-acc}
\vspace{-3mm}
\end{table}

\subsection{Baseline Models}


To our knowledge, the problem of generating scene layouts (with both \textit{thing} and \textit{stuff} categories, and multiple bounding boxes per category covering almost the entire image) from a label set is new and has not been studied. 
There is no existing model for this task so we adapt existing models that generate scene layout from more complex inputs \eg sentence, scene graphs.
Note that we choose the same distributions as in LayoutVAE for modeling the outputs (counts or bounding box information) to have a fair comparison between all the models.

\vspace{0.05in}
\noindent \textbf{Autoregressive MLP (AR-MLP).} This model is analogous to our proposed VAE based model, except that it has a multi-layer perceptron instead of a VAE. It runs autoregressively and takes as input the same conditioning information used in the VAE models. As in LayoutVAE, we have two separate sub-models -- \textit{CountMLP} for predicting count distribution given the corresponding conditioning information, and \textit{BBoxMLP} for predicting bounding box distribution given the corresponding conditioning information. 

\vspace{0.05in}
\noindent \textbf{BBoxLSTM.} This model is analogous to the bounding box predictor used in Hong~\etal\cite{hong2018}. It consists of an LSTM that takes in the label set embedding (as opposed to sentence embedding in \cite{hong2018}) at each step and predicts the label and bounding box distributions for bounding boxes in the image one by one. At each step, the LSTM output is first used to generate the label distribution. The bounding box distribution is then generated conditioned on the label. 

\vspace{0.05in}
\noindent \textbf{BLSTM.} We also use bidirectional LSTM in an analogous fashion as LayoutVAE. We have \textit{CountBLSTM} and \textit{BBoxBLSTM} to predict count distribution and bounding box distribution respectively, where we input the respective conditioning information at each step for the BLSTM. The conditioning information is similar as in the VAE models except that we do not give the pooled embedding of previous counts/bounding boxes information as we now have a bidirectional recurrent model. Note that having a bidirectional model can possibly alleviate the need to explicitly define the order in the label and the bounding box coordinates spaces. 

\vspace{0.05in}
\noindent \textbf{sg2im \cite{johnson2018}.} 
This model generates a scene layout based on a scene graph. 
We compare LayoutVAE with this model because label set (in this case, with multiple copies of labels as per the ground truth count of each label) can be seen as the simplest scene graph without any relations. 
We also explored another strategy where a scene graph is randomly generated for the label set, but we found that model performance worsened in that setting. For this experiment, we use the code and the pretrained model on COCO provided by the authors at \url{https://github.com/google/sg2im} to predict the bounding boxes.

\vspace{0.05in}
Note that these models are limited in their ability to model complex distributions and generate diverse layouts, whereas LayoutVAE can do so using the stochastic latent variable.

\begin{table}[t]
\centering
\begin{tabular}{lccccc}\toprule
\multirow{2}{*}{Model} & \multicolumn{2}{c}{Label Count} &  \multicolumn{2}{c}{BBox} \\
\cmidrule(lr){2-5}
& MNIST & COCO & MNIST & COCO \\
\midrule
AR-MLP & $1.246$ & $0.740$ & $5.83$ & $40.91$  \\
BBoxLSTM & - & - & $6.21$ & $42.42$ \\
BLSTM & $1.246$ & $0.732$ & $20.06$ & $52.84$ \\
sg2im \cite{johnson2018} & - & - & - & $214.03$ \\ 
LayoutVAE & $1.246$ & $\mathbf{0.562}$ & $\mathbf{-0.07}$ & $\mathbf{2.72}$ \\
\bottomrule
\end{tabular}
\vspace{-3mm}
\caption{\textbf{Comparison with baselines using likelihood metric.} Negative log-likelihood (lower is better) over the test set on MNIST-Layouts and COCO datasets. 
}
\label{tab:test-ll}
\vspace{-3mm}
\end{table}

\begin{table}[t]
\small
\centering
\begin{tabular}{cccc}
\toprule
\multirow{2}{*}{History} & \multirow{2}{*}{Context} & \multicolumn{2}{c}{NLL} \\
\cmidrule(lr){3-4}
& & CountVAE & BBoxVAE\\
\midrule
& & $0.592$ & $4.17$ \\ 
& \cmark & $0.570$ & $3.78$ \\
\cmark & & $0.581$ & $2.93$ \\
\cmark & \cmark &  $ \mathbf{0.562}$ & $\mathbf{2.72}$ \\
\bottomrule
\end{tabular}
\vspace{-3mm}
\caption{\textbf{Ablation study.} Test NLL results for CountVAE and BBoxVAE by prior sampling on COCO dataset. 
The history is the previous counts for CountVAE and the previous bounding box information for BBoxVAE.
The context is the label set for CountVAE and the label set with counts for BBoxVAE.
}
\label{tab:exp_ablation}
\vspace{-1mm}
\end{table}

\begin{table}
\centering
\begin{tabular}{ccccc}
\toprule
\textit{things} before \textit{stuffs} & \textit{stuffs} before \textit{things} & random \\
\midrule
$2.72$ & $\mathbf{2.71}$ & $3.22$ \\
\bottomrule
\end{tabular}
\vspace{-3mm}
\caption{\textbf{Analysis of label order.} NLL for BBoxVAE for different label set order on COCO dataset.}
\label{tab:order}
\vspace{-3mm}
\end{table}

\subsection{Quantitative Evaluation}
To compare the models, we compute average negative log-likelihood (NLL) of the test samples' count or bounding box coordinates under the respective predicted distributions. 
We compute Monte Carlo estimate of NLL for the VAE models by drawing 1000 samples from the conditional prior at each step of generation. 
LayoutVAE and the baselines are trained and evaluated by teacher forcing \ie we provide the ground truth context and count/bounding box history at each step of computation.

\vspace{0.05in}
\noindent \textbf{Comparison with baseline models.}
\autoref{tab:test-acc}(accuracy metrics) and \autoref{tab:test-ll}(likelihood) show count and bounding box generation performances for all the models.
First, we observe that LayoutVAE model is significantly better than all the baselines. 
In particular the large improvement with respect to the autoregrssive MLP baseline shows the relevance of stochastic model for this task. 
It is also interesting to note that autoregressive MLP model performs better that the recurrent models.
Finally, we observe that sg2im model \cite{johnson2018} is not able to predict accurate bounding boxes without a scene graph with relationships. LayoutVAE and the baselines show similar performance for count prediction in MNIST-Layouts because estimating count distribution over the 4 labels of MNIST-Layouts is much easier than in the real world data from COCO.

\vspace{0.05in}
\noindent \textbf{Ablation study.}
In \autoref{tab:exp_ablation}, we analyze the importance of context representation and history in the conditioning information.
We observe that both history and context are useful because they increase the log-likelihood but their effects vary for count and bounding box models. The context is more critical than the history for the CountVAE whereas the history is more critical than the context for the BBoxVAE. 
Despite these different behaviours, we note that history and context are complementary and increase the performances for both CountVAE and BBoxVAE.
This experiment validates the importance of both context and history in the conditioning information for both CountVAE and BBoxVAE.

\vspace{0.05in}
\noindent \textbf{Analysis of the label set order.}
We performed experiments by changing the order of labels in the label set.
We consider three variants for this experiments --- a fixed order with all the \textit{things} before \textit{stuffs} (default choice for all the other experiments), reverse order of the above with all the \textit{stuffs} before \textit{things}, and finally we randomly assigned the order of labels for each image. 
\autoref{tab:order} shows the results of the experiment on BBoxVAE. We found that fixed order (\textit{things} first or \textit{stuffs} first) gave similar results, while randomizing the order of labels across images resulted in a significant reduction in performance.

\begin{figure}[t]
\centering
\begin{tabular}{cccccc}
\hspace{-4mm}
\pdftooltip{\includegraphics[width = 1.1in]{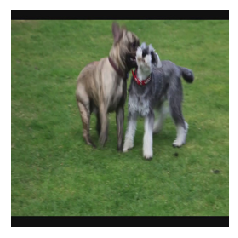}}{Image from COCO dataset, originally from Flickr.com and available at http://farm4.staticflickr.com/3353/3424594672_333e3ee522_z.jpg, licensed under CC BY 2.0 (https://creativecommons.org/licenses/by/2.0/)} &
\hspace{-5mm}
\includegraphics[width = 1.1in]{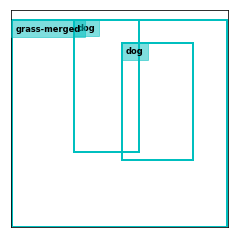} &
\hspace{-5mm}
\includegraphics[width = 1.1in]{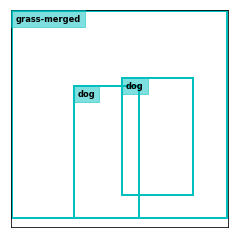} 

\vspace{-2mm}
\\ 
\hspace{-4mm} original image & \hspace{-5mm} original layout & \hspace{-5mm} flipped layout &
\end{tabular}
\vspace{-2mm}
\caption{Example where likelihood under LayoutVAE increases when flipped upside down. 
We can see that the flipped layout is equally plausible for this example.}
\label{fig:coco_flipped}
\vspace{-1mm}
\end{figure}

\begin{figure}
\begin{tabular}{ccccc}
\hspace{-2mm} 
\includegraphics[width=0.6in]{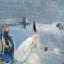}
& \hspace{-4mm} \includegraphics[width=0.6in]{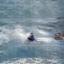}
& \hspace{-4mm} \includegraphics[width=0.6in]{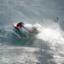}
& \hspace{-4mm} \includegraphics[width=0.6in]{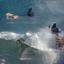}
& \hspace{-4mm} \includegraphics[width=0.6in]{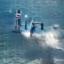}
\\
\hspace{-2mm}
\includegraphics[width=0.6in]{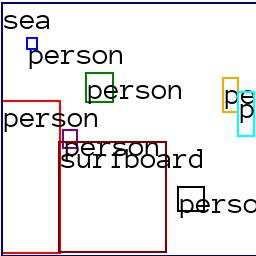}
& \hspace{-4mm} \includegraphics[width=0.6in]{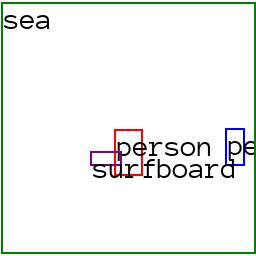}
& \hspace{-4mm} \includegraphics[width=0.6in]{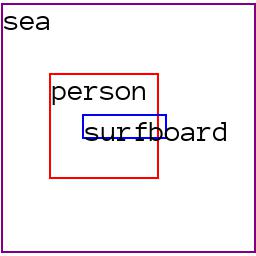}
& \hspace{-4mm} \includegraphics[width=0.6in]{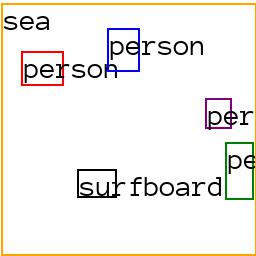}
& \hspace{-4mm} \includegraphics[width=0.6in]{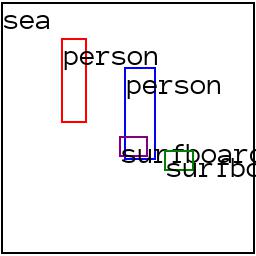}
\end{tabular}
\vspace{-2mm}
\caption{\textbf{Image Generation from a label set.} We show diverse realistic layouts generated by LayoutVAE for the input label set \textit{person}, \textit{surfboard} and \textit{sea}. We generate images from the layouts by using the generative model provided by Zhao~\etal~\cite{zhao2018}}
\label{fig:coco_gen}
\vspace{-3mm}
\end{figure}

\begin{figure*}[ht]
\centering
\begin{subfigure}[b]{0.475\textwidth}
\centering
\includegraphics[width=\textwidth]{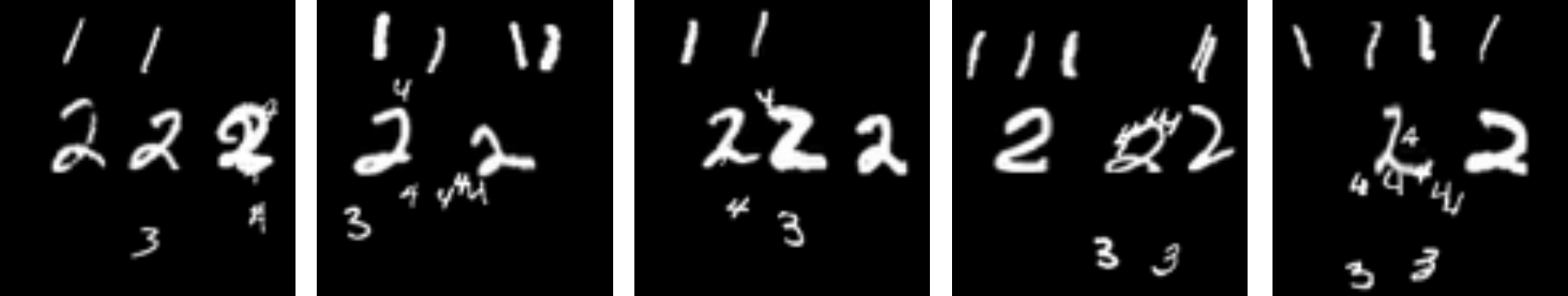}
\vspace{-6mm}\caption[]%
{{\small Input label set $\{1, 2, 3, 4\}$}}    
\end{subfigure}
\quad
\begin{subfigure}[b]{0.475\textwidth}  
\centering 
\includegraphics[width=\textwidth]{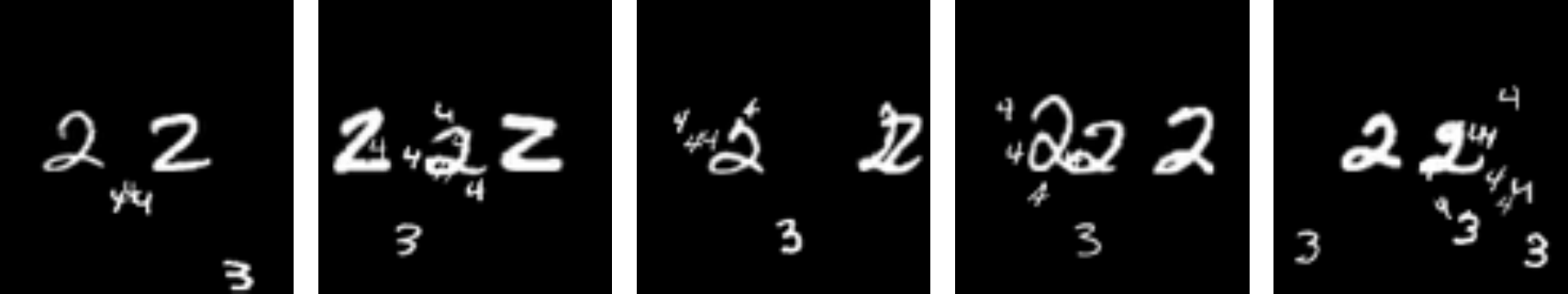}
\vspace{-6mm}\caption[]%
{{\small Input label set $\{2, 3, 4\}$}}    
\end{subfigure}
\begin{subfigure}[b]{0.475\textwidth}   
\centering 
\includegraphics[width=\textwidth]{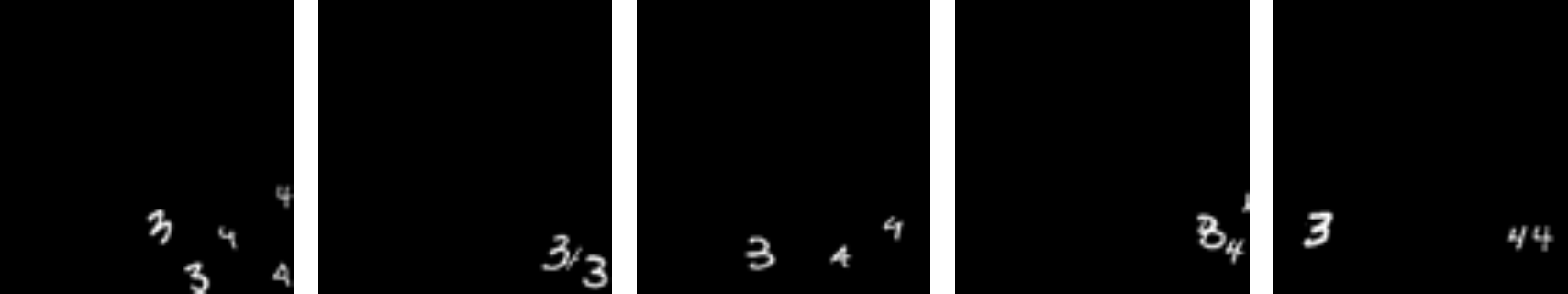}
\vspace{-6mm}\caption[]%
{{\small Input label set $\{3, 4\}$}}    
\end{subfigure}
\quad
\begin{subfigure}[b]{0.475\textwidth}   
\centering 
\includegraphics[width=\textwidth]{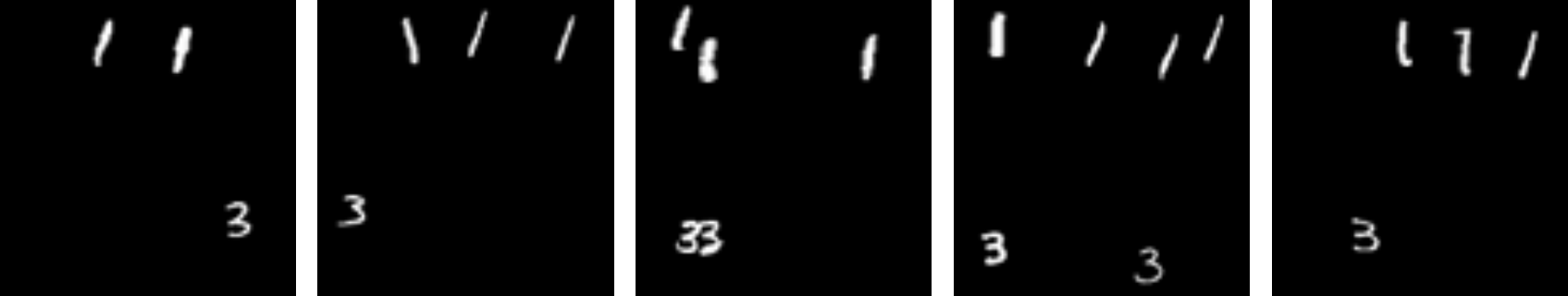}
\vspace{-6mm}\caption[]%
{{\small Input label set $\{1, 3\}$}}    
\end{subfigure}
\vspace{-3mm}\caption{\textbf{Stochastic layout generation for MNIST-Layouts.} We show 5 sample layouts generated by LayoutVAE for each input set of labels. We fill in the generated bounding boxes with randomly sampled MNIST digits of appropriate labels. We can see that LayoutVAE generates bounding boxes following all the rules that we defined for MNIST-Layouts (e.g.\ large 2s in the middle, small 3s at the bottom etc.\ with the appropriate count values). The complete set of rules is given in \autoref{sec:supp_mnist_layout} in the supplementary material.} 
\label{fig:mnist_gen}
\vspace{-3mm}
\end{figure*}

\begin{figure*}[t]
\centering
\pdftooltip{\begin{tabular}{ccccccc}
\hspace{-4mm} \includegraphics[width=1.2in]{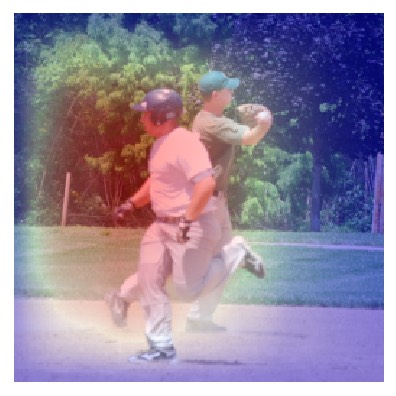}
& \hspace{-6.5mm} \includegraphics[width=1.2in]{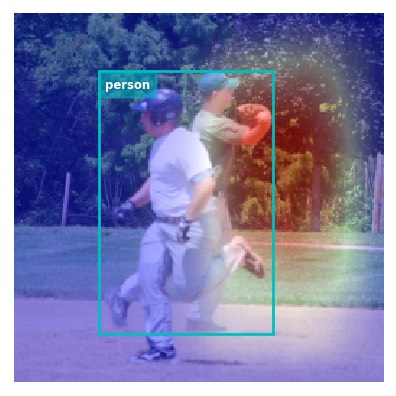}
& \hspace{-6.5mm} \includegraphics[width=1.2in]{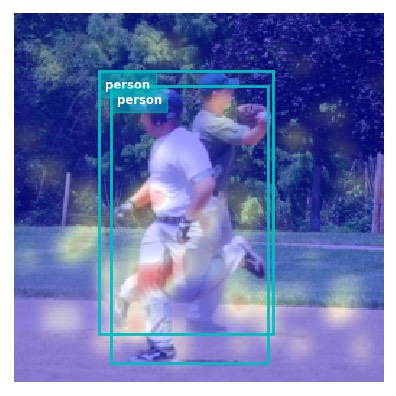}
& \hspace{-6.5mm} \includegraphics[width=1.2in]{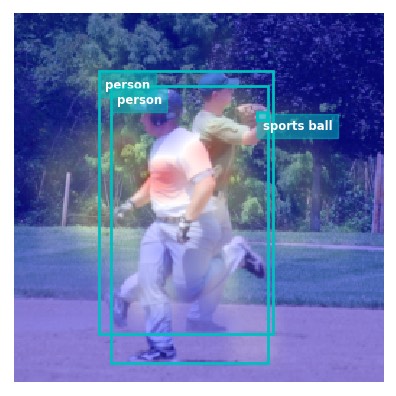}
& \hspace{-6.5mm} \includegraphics[width=1.2in]{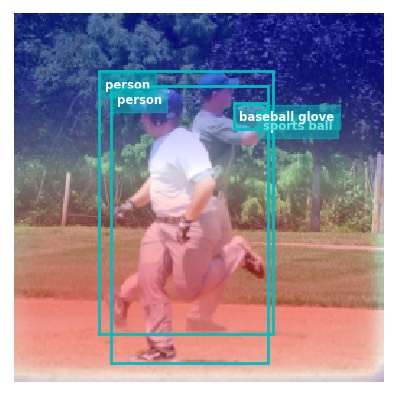}
& \hspace{-6.5mm} \includegraphics[width=1.2in]{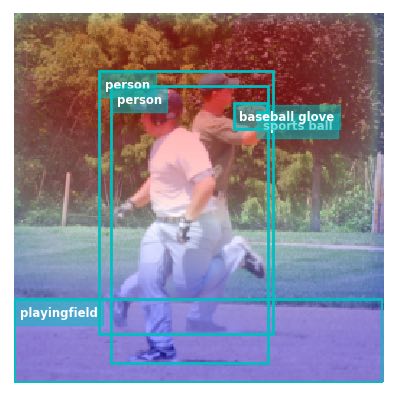}
\vspace{-2.5mm}
\\
\hspace{-4mm}
\includegraphics[width=1.2in]{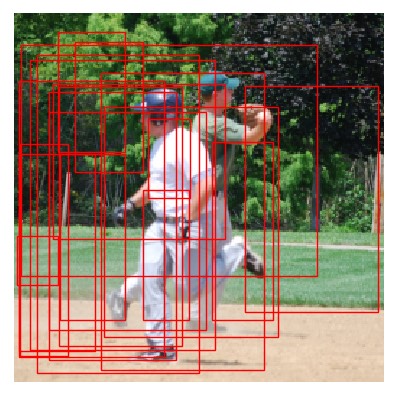}
& \hspace{-6.5mm} \includegraphics[width=1.2in]{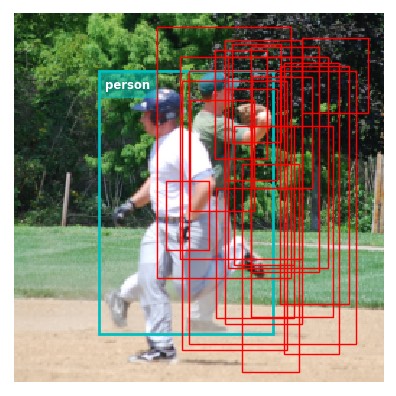}
& \hspace{-6.5mm} \includegraphics[width=1.2in]{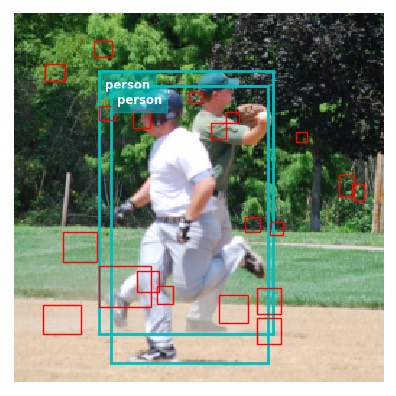}
& \hspace{-6.5mm} \includegraphics[width=1.2in]{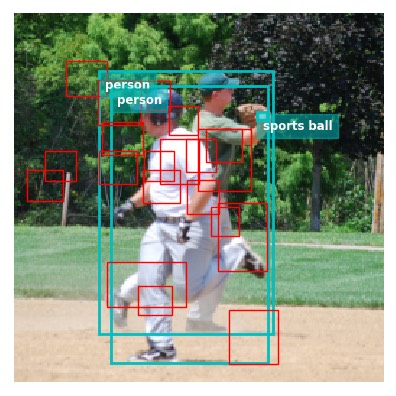}
& \hspace{-6.5mm} \includegraphics[width=1.2in]{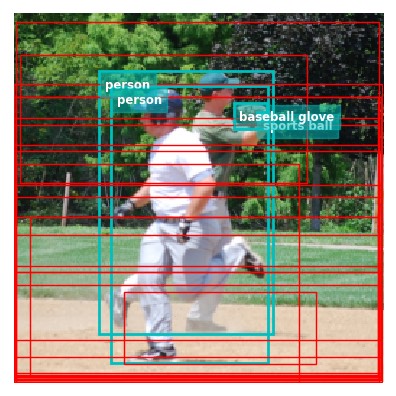}
& \hspace{-6.5mm} \includegraphics[width=1.2in]{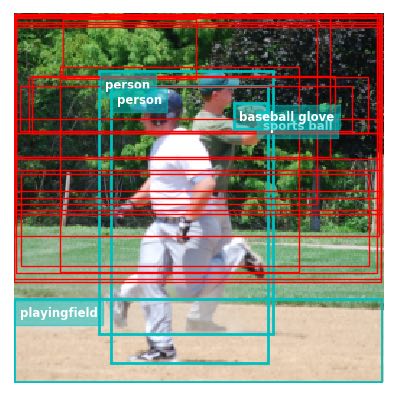}
\vspace{-1.5mm}
\\
\hspace{-4mm} person & \hspace{-6.5mm} person & \hspace{-6.5mm} sports ball & \hspace{-6.5mm} baseball glove & \hspace{-6.5mm} playingfield & \hspace{-6.5mm} tree-merged 
\end{tabular}}{Image from COCO dataset, originally from Flickr.com and available at http://farm9.staticflickr.com/8447/7805810128_605424213d_z.jpg, licensed under CC BY 2.0 (https://creativecommons.org/licenses/by/2.0/)}
\vspace{-3mm}
\caption{\textbf{Bounding box prediction using LayoutVAE.} We show the steps of bounding box generation for a test set sample. At each step, we obtain diverse bounding box predictions for the input label (written in subcaptions) given the ground truth layout up to the previous step (shown in cyan bounding boxes). We show bounding box heatmap using 100 samples from the prior distribution (top row for each step, red means high probability), and  bounding boxes using $20$ samples from the prior distribution. The predicted layouts are overlaid on the test image for clarity.
}
\label{fig:coco_bb}
\vspace{-3mm}
\end{figure*}

\vspace{0.05in}
\noindent \textbf{Detecting unlikely layouts in COCO.} 
To test the ability of the model to differentiate between plausible and unlikely layouts, we perform an experiment where we flip the image layouts in the test set upside down, and evaluate the NLL (in this case, by importance sampling using $1000$ samples from the posterior) of both the normal layout and the flipped layout under BBoxVAE. 
We found that the NLL got worse for $92.58\%$ of image layouts in the test set when flipped upside down. 
The average NLL for original layouts is $2.26$ while that for the flipped layouts is $4.33$. 
We note that there are instances where flipping does not necessarily lead to an anomalous layout, which explains why likelihood worsened only for 92.58\% of the test set layouts.
\autoref{fig:coco_flipped} shows such an example where flipping led to an increased likelihood under the model. Additional results are shown in \autoref{sec:supp_unlikely_layouts} in the supplementary material.

\subsection{Qualitative Evaluation}
\label{sec:qual}

\noindent \textbf{Scene layout samples.} 
We present qualitative results for diverse layout generation on COCO(\autoref{fig:coco_gen}) datasets and MNIST-Layouts(\autoref{fig:mnist_gen}). 
The advantages of modeling scene layout generation using a probabilistic model is evident from these results: LayoutVAE learns the rules of object layout in the scene and is capable of generating diverse layouts with different counts of objects. More examples can be found in \autoref{sec:sup_layout_generation} in the supplementary material.

\vspace{0.05in}
\noindent \textbf{Per step bounding box prediction for COCO.} 
\autoref{fig:coco_bb} shows steps of bounding box generation for a test set of labels. 
For each step, we show the probability map and 20 plausible bounding boxes.
We observe that given some bounding boxes, the model is able to predict where the next object could be in the image. 
More examples for bounding box generation are provided in \autoref{sec:supp_bbox_generation} in the supplementary material.

\section{Conclusion}

We propose LayoutVAE for generating stochastic scene layouts given a label set as input. 
It comprises of two components to model the distribution of spatial and count relationships between objects in a scene. 
We compared it with other existing methods or analogues thereof, and showed significant performance improvement. Qualitative visualizations show that LayoutVAE can learn intrinsic relationships between objects in real world scenes.  Furthermore, LayoutVAE can be used to detect abnormal layouts.


{\small
\bibliographystyle{ieee_fullname}
\bibliography{main}
}
\clearpage


\onecolumn
\appendix

\appendix
\addtocounter{figure}{7}
\addtocounter{equation}{13}
\addtocounter{table}{4}

\section{Graphical Model}
\label{app:graphical_model}
\begin{figure}[h]
\centering
\includegraphics[width=0.8\linewidth]{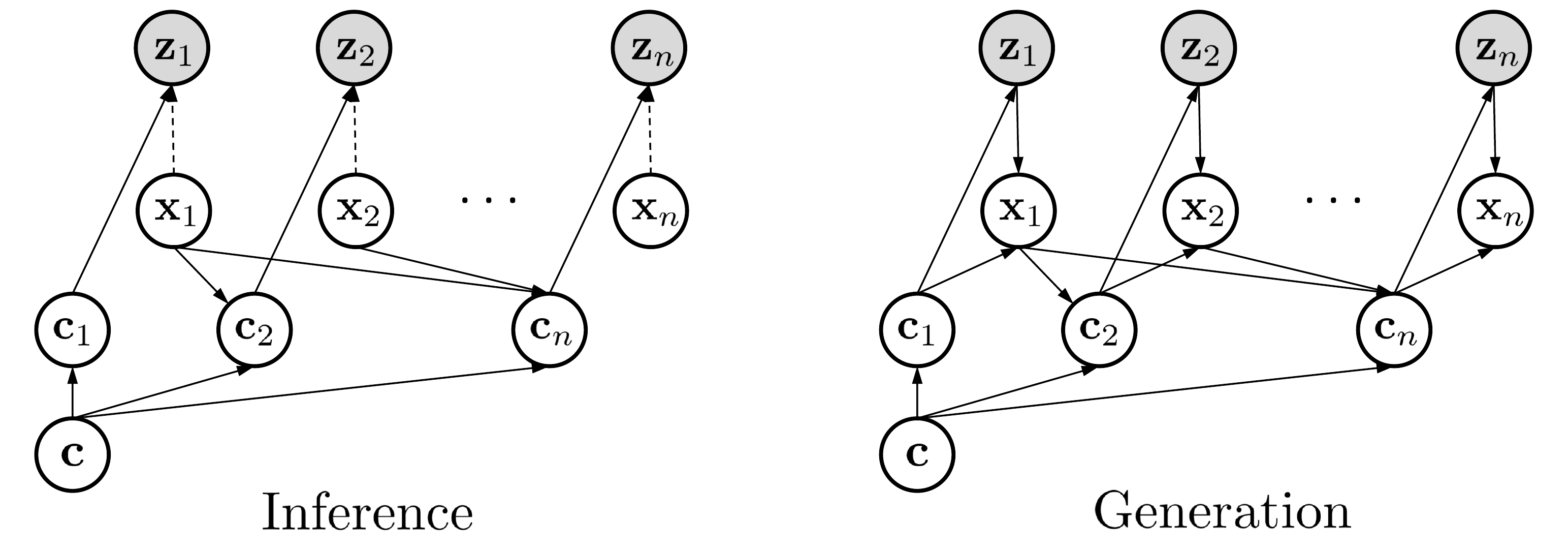}
\caption{\textbf{Graphical model for LayoutVAE.} $\mathbf{c}$ denotes context (label set for CountVAE, label set with counts for BBoxVAE) and $\mathbf{c}_k$ denotes the conditioning information for the CVAE. Dashed arrow denotes inference of the approximate posterior.}
\label{fig:graphical_model}
\end{figure}

For each of CountVAE and BBoxVAE, we have a conditional VAE that is autoregressive (\autoref{fig:graphical_model}), where the conditioning variable contains all the information required for each step of generation. By explicitly designing the conditioning variable this way, we forgo using past latent codes for generation.

\section{Model Architecture}
\subsection{CountVAE}
\label{app:countVAE}

\begin{figure*}[b]
\includegraphics[width=\textwidth]{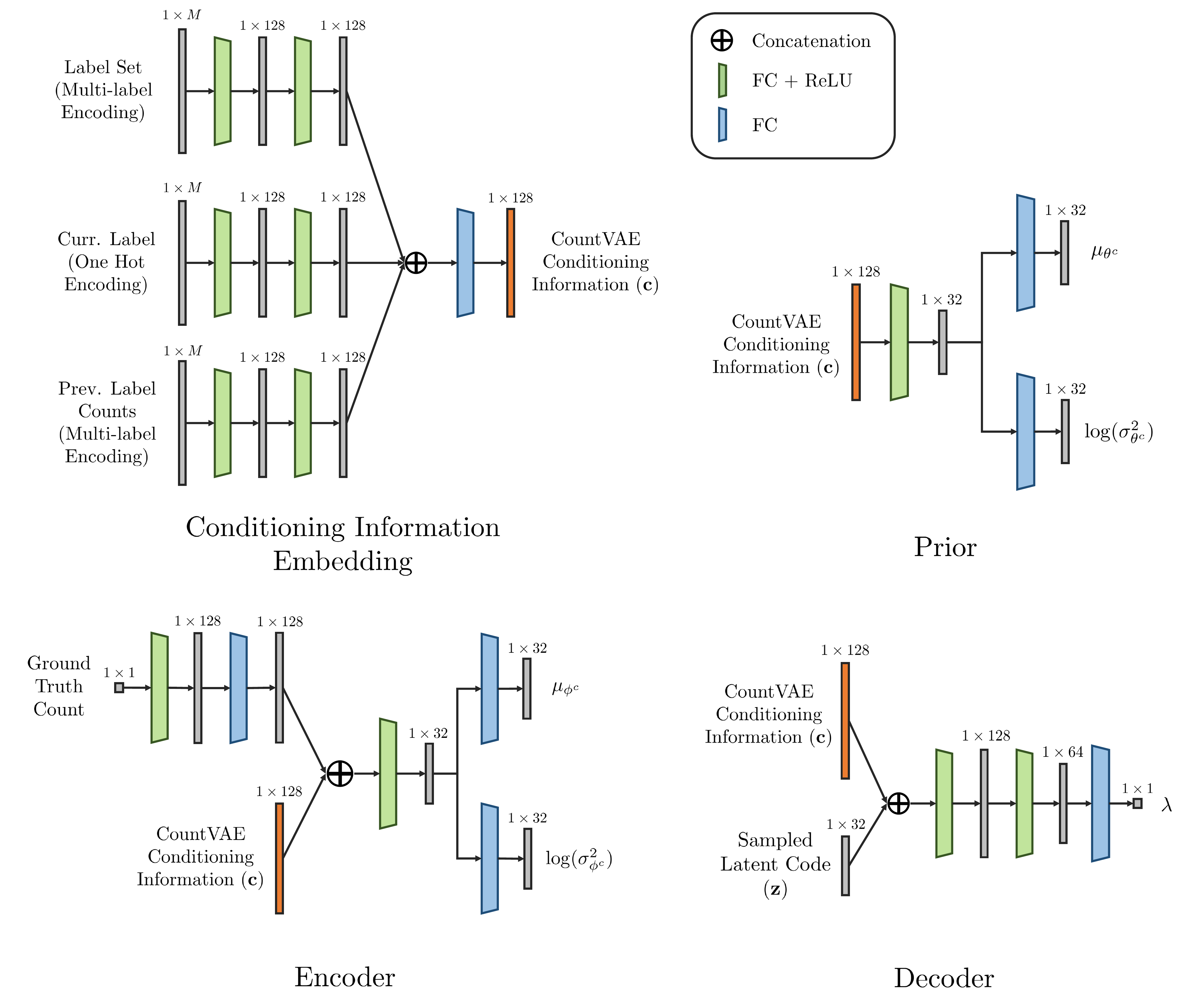}
\caption{\textbf{CountVAE Architecture.}
}
\label{fig:CountVAE}
\end{figure*}

We represent the label set $\lset$ by a multi-label vector $\lsetvec \in \{0, 1\}^M$, where $\lsetvec_k = 1$ (resp. 0) means the $k$-th category is present (resp. absent).
For each step of CountVAE, the current label $k$ is represented by a one-hot vector denoted as $\lab^k \in \{0, 1\}^M$.
We represent the count information $\co_k$ of category $k$ as a $M$ dimensional one-hot vector $\mathbf{y}^k$ with the non-zero location filled with the count value. 
This representation of count captures its label information as well. 
We perform pooling over a set of previously predicted label counts by summing up the vectors.
The conditioning input is 
\begin{align}
    \mathbf{c}^c_k= \fc\left( \left[\mlp(\lsetvec), \mlp(\lab^k), \mlp \left(\sum_{m<k} \mathbf{y}^k \right) \right] \right)
\end{align}
where $[\cdot,\cdot]$ denotes concatenation, $\fc$ a fully connected layer and $\mlp$ a generic multi-layer perceptron. \autoref{fig:CountVAE} shows the architecture in detail.

\subsection{BBoxVAE}
\label{app:BBoxVAE}

\begin{figure*}[b]
\includegraphics[width=\textwidth]{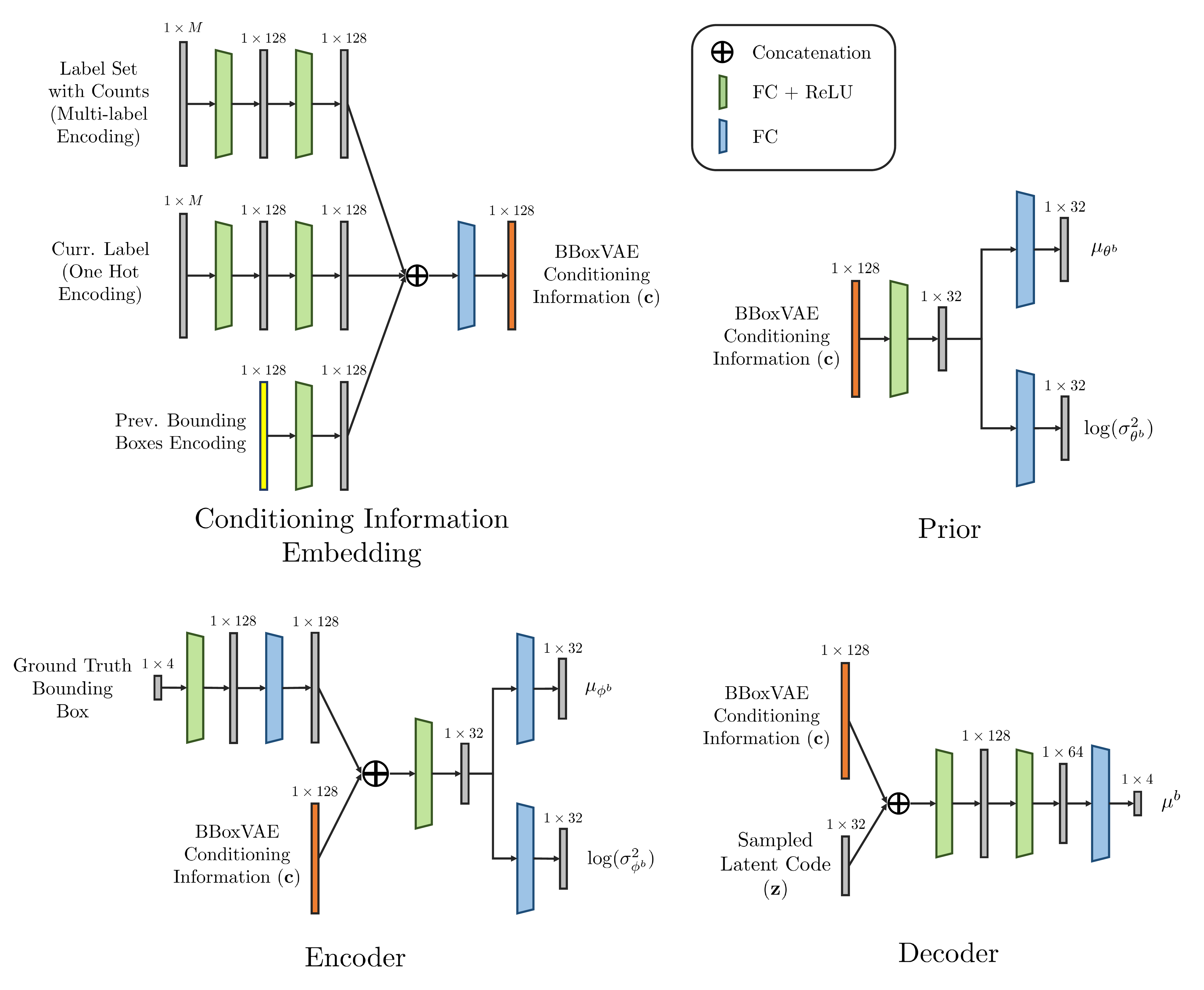}
\caption{\textbf{BBoxVAE Architecture.}
}
\label{fig:BBoxVAE}
\end{figure*}

\begin{figure*}
\centering
\includegraphics[width=0.8\textwidth]{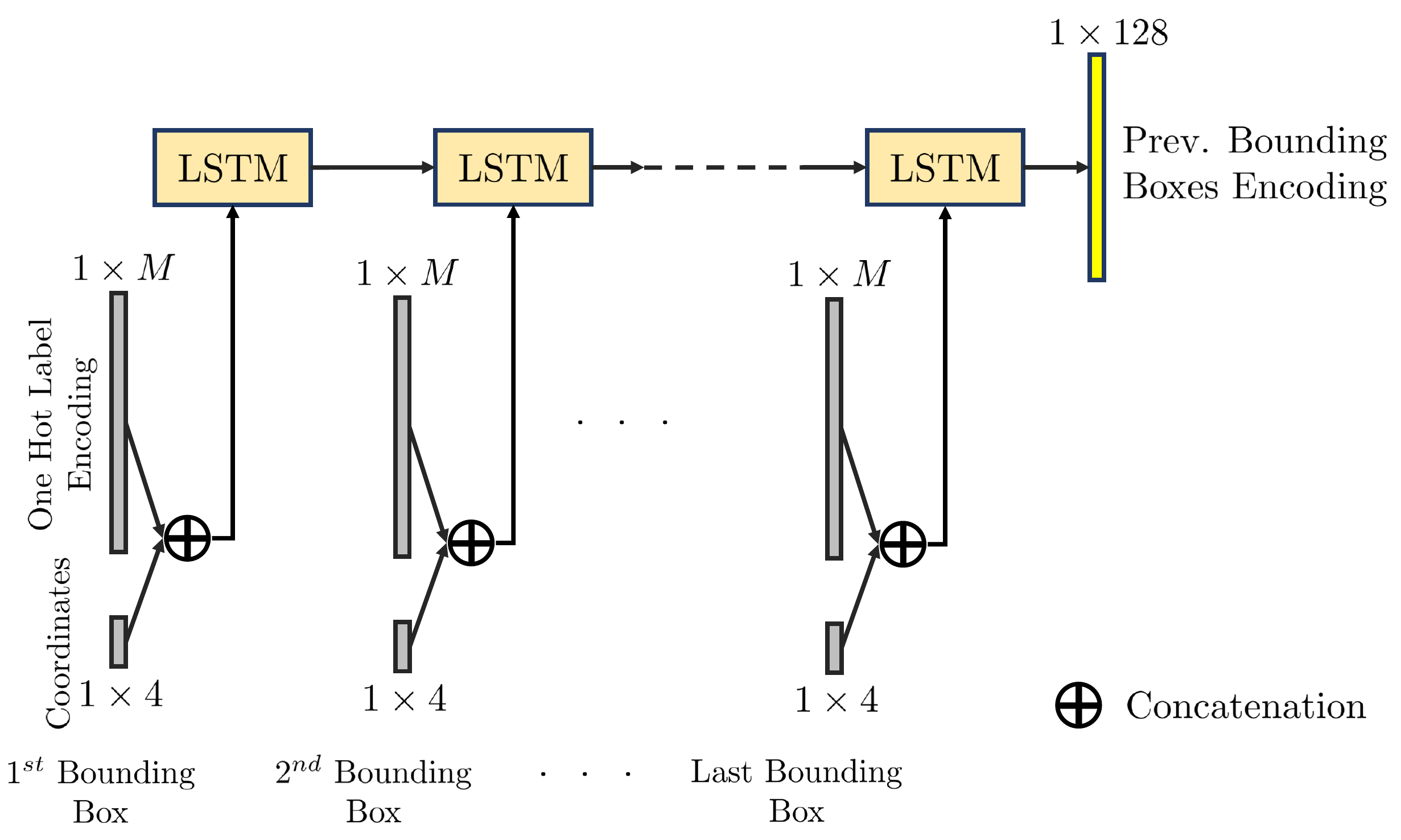}
\caption{\textbf{LSTM pooling for bounding boxes.} We use an LSTM to pool the set of previously predicted bounding boxes to be used in the conditioning information for BBoxVAE.}
\label{fig:BBox_pooling}
\end{figure*}

We represent label and count information pair for each category $\{\co_m : m \in \lset\}$ as $\mathbf{y}^m$ using the same strategy as in CountVAE. Pooled representation of the label set along with counts is obtained by summing up these vectors to obtain a multi-label vector.
For each step of BBoxVAE, the current label $k$ is represented by a one-hot vector denoted as $\lab^k$.
We use LSTM for pooling previously predicted bounding boxes $\bboxes_{k,j}^{prev}$. 
We represent each bounding box as a vector of size 4, and we concatenate $M$ dimensional label vector to add label information to it. 
We pass $M+4$ dimensional vectors of successive bounding boxes through an LSTM and use the final step output as the pooled representation. \autoref{fig:BBox_pooling} shows the pooling operation, and \autoref{fig:BBoxVAE} shows the detailed architecture for each module in BBoxVAE. The conditioning input is
\begin{align}
\mathbf{c}^b_{k,j}= \fc \left( \left[ \mlp \left(\sum_{m \in \lset} \mathbf{y}^m \right), \mlp(\lab^k), \mlp(\bboxes_{k,j}^{prev}) \right] \right).
\end{align}
BBoxVAE predicts the mean for the quadrivariate Gaussian (\autoref{eq:gaussian}), while covariance is assumed to be a diagonal matrix with each value of standard deviation equal to $0.02$.

\clearpage
\section{Experiments}

\subsection{MNIST-Layouts dataset}
\label{sec:supp_mnist_layout}

\autoref{tab:rules} shows the rules used to generate the dataset.
To generate the layouts, we adapted the code provided at \url{https://github.com/aakhundov/tf-attend-infer-repeat}.

\begin{table}[H]
\centering
\begin{tabular}{@{}cccc@{}}\toprule
Label & Count & Location & Size
\\ \midrule
1 & 3,4 & top  & medium \\[0.1cm]
2 & 2,3 & middle  & large \\[0.1cm]
3 & 1,2 & bottom  & small-medium \\[0.2cm]
4 (2 present) & count(2)+3,6 & 
around a 2 & small \\[0.2cm]
4 (2 absent) & 2 & bottom-right & small \\
\bottomrule
\end{tabular}
\vspace{0.05in}
\caption{\textbf{Rules for generating MNIST-Layouts dataset.} Given a label set, we use uniform distribution over the possible count values to generate count. We then sample over a uniform distribution over the location and size ranges (precise details skipped in the table for brevity) to generate bounding boxes for each label instance. When label $4$ is present in the input label along with label $2$ ($4^{th}$ row in the table), we randomly choose an instance of 2 and place all the 4s around that.}
\label{tab:rules}
\end{table}

\subsection{Analysis of latent code size}

In \autoref{tab:latent}, we analyze the dimension of the latent space for both CountVAE and BBoxVAE. 
For each dimension we report the NLL performance on COCO dataset.
We observe that both models have good performance when the latent code size is between 32 and 128, and the models are not sensitive to this hyperparameter. 
Increasing the latent space beyond 128 does not improve the performances.

\begin{table}[H]
\centering
\begin{tabular}{@{}ccc@{}}\toprule
Latent Code Size & CountVAE &  BBoxVAE 
\\ \midrule
$2$ & $0.569$ & $4.13$  \\
$4$ & $0.568$ & $3.11$ \\
$8$ & $0.565$ & $2.91$ \\
$16$ & $0.564$ & $2.72$ \\ 
$32$ & $\mathbf{0.562}$ & $2.72$ \\
$64$ & $0.563$ & $\mathbf{2.69}$ \\
$128$ & $\mathbf{0.562}$ & $2.70$ \\
\bottomrule
\end{tabular}
\vspace{0.05in}
\caption{\textbf{Effect of latent code size.} Average NLL over COCO test set for CountVAE and BBoxVAE while varying the size of the latent code. 
}
\label{tab:latent}
\end{table}

\clearpage
\subsection{Detecting unlikely layouts}
\label{sec:supp_unlikely_layouts}
\begin{figure*}[h]
\centering
\begin{tabular}{cccccc}
\hspace{-4mm}
\pdftooltip{\includegraphics[width = 1.1in]{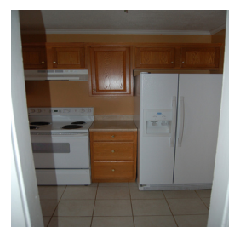}}
{Image from COCO dataset, originally from Flickr.com and available at http://farm5.staticflickr.com/4031/4430589002_88e591d3e9_z.jpg, licensed under CC BY-ND 2.0 (https://creativecommons.org/licenses/by-nd/2.0/)}
 &
\hspace{-4mm}
\includegraphics[width = 1.1in]{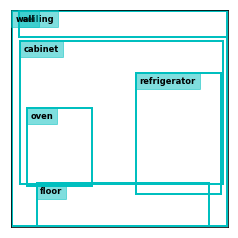} &
\hspace{-4mm}
\includegraphics[width = 1.1in]{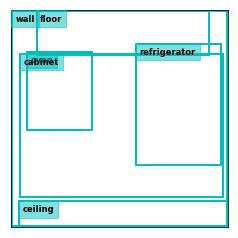} &
\hspace{-4mm}
\pdftooltip{\includegraphics[width = 1.1in]{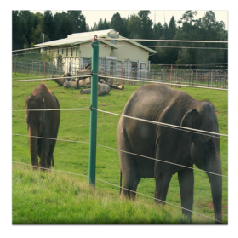}}
{Image from COCO dataset, originally from Flickr.com and available at http://farm5.staticflickr.com/4145/5036782631_1ccbc49c38_z.jpg, licensed under CC BY-NC-SA 2.0 (https://creativecommons.org/licenses/by-nc-sa/2.0/)}
&
\hspace{-4mm}
\includegraphics[width = 1.1in]{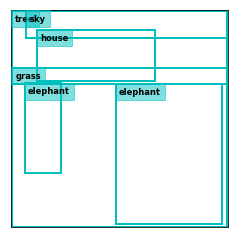} &
\hspace{-4mm}
\includegraphics[width = 1.1in]{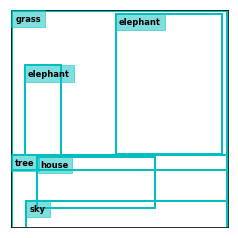}
\\ 
\hspace{-4mm}  & \hspace{-4mm} NLL $=1.81$  & \hspace{-4mm} NLL $=11.69$ &
\hspace{-4mm}  & \hspace{-4mm} NLL $=1.95$ & \hspace{-4mm} NLL $=11.15$
\\
\hspace{-4mm}
\pdftooltip{\includegraphics[width = 1.1in]{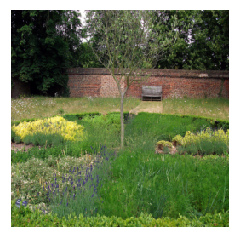}}
{Image from COCO dataset, originally from Flickr.com and available at http://farm6.staticflickr.com/5226/5791374148_f4e99e57a9_z.jpg, licensed under CC BY-NC-ND 2.0 (https://creativecommons.org/licenses/by-nc-nd/2.0/)}
&
\hspace{-4mm}
\includegraphics[width = 1.1in]{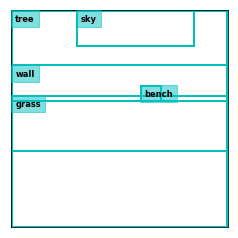} &
\hspace{-4mm}
\includegraphics[width = 1.1in]{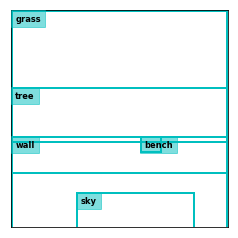} &
\hspace{-4mm}
\pdftooltip{\includegraphics[width = 1.1in]{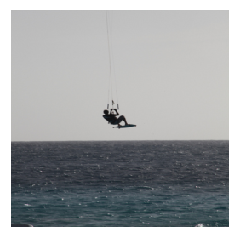}}
{Image from COCO dataset, originally from Flickr.com and available at http://farm8.staticflickr.com/7279/7056003099_2652720fb3_z.jpg, licensed under CC BY-NC-ND 2.0 (https://creativecommons.org/licenses/by-nc-nd/2.0/)} 
&
\hspace{-4mm}
\includegraphics[width = 1.1in]{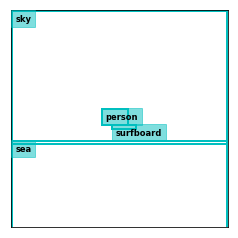} &
\hspace{-4mm}
\includegraphics[width = 1.1in]{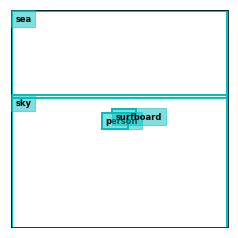}
\\ 
\hspace{-4mm}  & \hspace{-4mm} NLL $=2.15$ & \hspace{-4mm} NLL $=11.26$ &
\hspace{-4mm}  & \hspace{-4mm} NLL $=0.71$ & \hspace{-4mm} NLL $=9.07$
\vspace{3mm}
\\ 
\hspace{-4mm} image & \hspace{-4mm} image layout & \hspace{-4mm} flipped layout &
\hspace{-4mm} image & \hspace{-4mm} image layout & \hspace{-4mm} flipped layout
\end{tabular}
\caption{\textbf{Some examples where likelihood under BBoxVAE decreases when flipped upside down.} We show the test image, layout for the image and the flipped layout. Negative log likelihood(NLL) of the layout under BBoxVAE is shown along with each layout. We can see that the flipped layout is highly unlikely in these examples.}
\label{fig:supp_coco_flipped_worse}
\end{figure*}

\begin{figure*}[h]
\centering
\begin{tabular}{cccccc}
\hspace{-4mm}
\pdftooltip{\includegraphics[width = 1.1in]{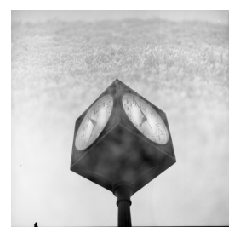}}
{Image from COCO dataset, originally from Flickr.com and available at http://farm9.staticflickr.com/8177/8011171833_87909e07b4_z.jpg, licensed under CC BY-NC-SA 2.0 (https://creativecommons.org/licenses/by-nc-sa/2.0/)}
&
\hspace{-4mm}
\includegraphics[width = 1.1in]{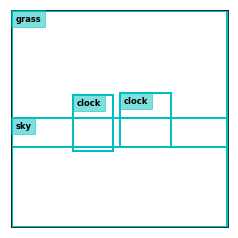} &
\hspace{-4mm}
\includegraphics[width = 1.1in]{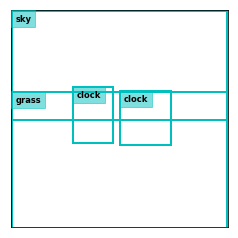} &
\hspace{-4mm}
\pdftooltip{\includegraphics[width = 1.1in]{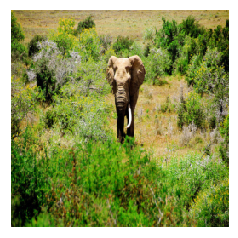}}
{Image from COCO dataset, originally from Flickr.com and available at http://farm1.staticflickr.com/176/369369838_f0c65e2c6e_z.jpg, licensed under CC BY-NC-ND 2.0 (https://creativecommons.org/licenses/by-nc-nd/2.0/)}
&
\hspace{-4mm}
\includegraphics[width = 1.1in]{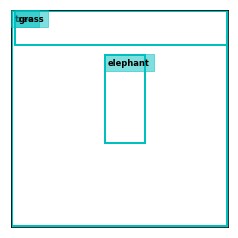} &
\hspace{-4mm}
\includegraphics[width = 1.1in]{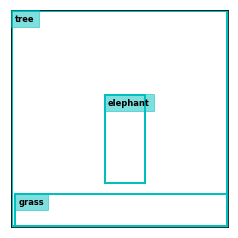}
\\ 
\hspace{-4mm}  & \hspace{-4mm} NLL $=6.15$  & \hspace{-4mm} NLL $=1.89$ &
\hspace{-4mm}  & \hspace{-4mm} NLL $=3.47$ & \hspace{-4mm} NLL $=2.16$
\\
\hspace{-4mm}
\pdftooltip{\includegraphics[width = 1.1in]{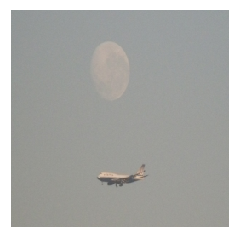}}
{Image from COCO dataset, originally from Flickr.com and available at http://farm7.staticflickr.com/6105/6349058853_36de894aa1_z.jpg, licensed under CC BY 2.0 (https://creativecommons.org/licenses/by/2.0/)}
&
\hspace{-4mm}
\includegraphics[width = 1.1in]{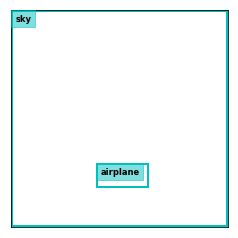} &
\hspace{-4mm}
\includegraphics[width = 1.1in]{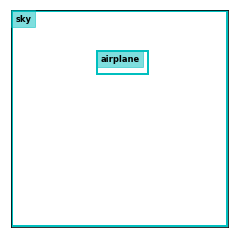} &
\hspace{-4mm}
\pdftooltip{\includegraphics[width = 1.1in]{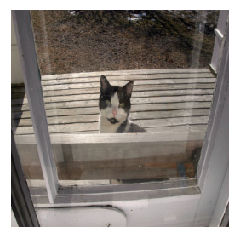}}
{Image from COCO dataset, originally from Flickr.com and available at http://farm3.staticflickr.com/2404/2240113245_411e0ed8ef_z.jpg, licensed under CC BY-NC-ND 2.0 (https://creativecommons.org/licenses/by-nc-nd/2.0/)}
&
\hspace{-4mm}
\includegraphics[width = 1.1in]{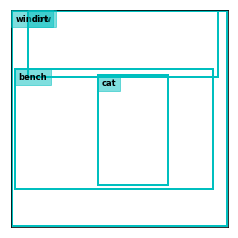} &
\hspace{-4mm}
\includegraphics[width = 1.1in]{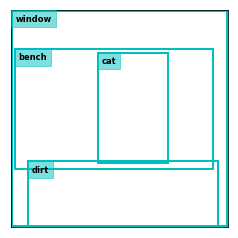}
\\ 
\hspace{-4mm}  & \hspace{-4mm} NLL $=2.19$ & \hspace{-4mm} NLL $=0.90$ &
\hspace{-4mm}  & \hspace{-4mm} NLL $=3.72$ & \hspace{-4mm} NLL $=2.62$
\vspace{3mm}
\\ 
\hspace{-4mm} image & \hspace{-4mm} image layout & \hspace{-4mm} flipped layout &
\hspace{-4mm} image & \hspace{-4mm} image layout & \hspace{-4mm} flipped layout
\end{tabular}
\caption{\textbf{Some examples where likelihood under BBoxVAE increases when flipped upside down.} We show the test image, layout for the image and the flipped layout. Negative log likelihood(NLL) of the layout under BBoxVAE is shown along with each layout. We can see that the flipped layout is equally or sometimes more plausible in these examples.}
\label{fig:supp_coco_flipped_better}
\end{figure*}

In this section, we present more examples from our experiment on detecting unlikely layouts by flipping the original layout and computing likelihood under LayoutVAE. In \autoref{fig:supp_coco_flipped_worse}, we present some typical examples where likelihood under LayoutVAE (BBoxVAE, to be precise, since CountVAE gives the same result for original and flipped layouts as the label counts remain the same) decreases when flipped upside down. This behaviour was observed for $92.58\%$ samples in the test set. In \autoref{fig:supp_coco_flipped_better}, we present some examples of unusual layouts where likelihood under LayoutVAE increases when flipped upside down.


\subsection{Examples for Layout Generation}
\label{sec:sup_layout_generation}

\begin{figure*}[h]
\centering
\begin{tabular}{ccccc}
\fbox{\includegraphics[width=1in]{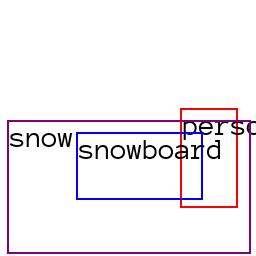}}
&\fbox{\includegraphics[width=1in]{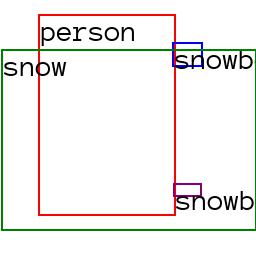}}
& \fbox{\includegraphics[width=1in]{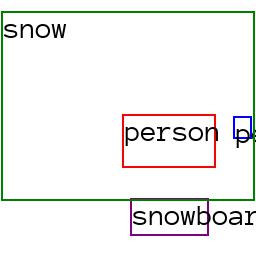}}
& \fbox{\includegraphics[width=1in]{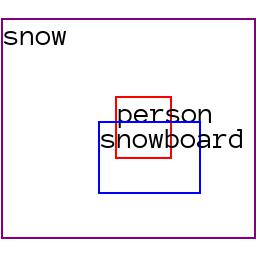}}
& \fbox{\includegraphics[width=1in]{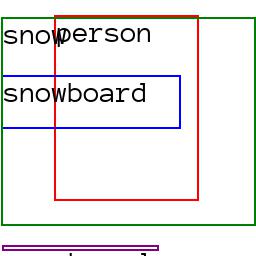}}
\\
\multicolumn{5}{c}{ \{person, snow, snowboard\} }
\vspace{3mm}
\\
\fbox{\includegraphics[width=1in]{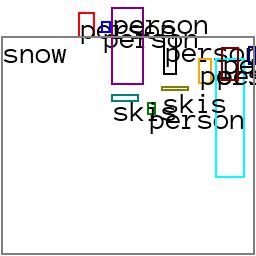}}
&\fbox{\includegraphics[width=1in]{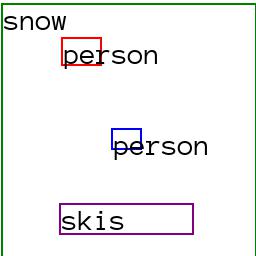}}
& \fbox{\includegraphics[width=1in]{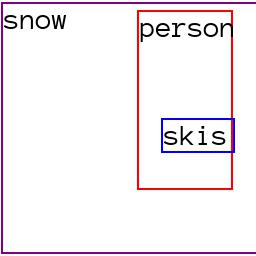}}
& \fbox{\includegraphics[width=1in]{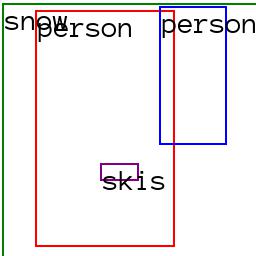}}
& \fbox{\includegraphics[width=1in]{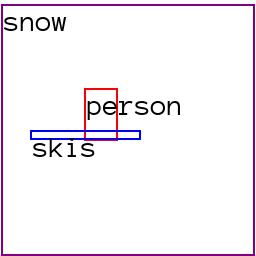}}
\\
\multicolumn{5}{c}{ \{person, snow, skis\} }
\vspace{3mm}
\\
\fbox{\includegraphics[width=1in]{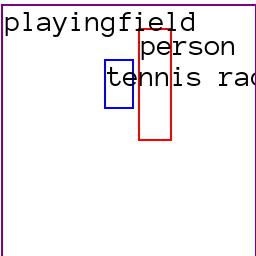}}
&\fbox{\includegraphics[width=1in]{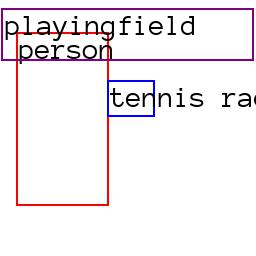}}
& \fbox{\includegraphics[width=1in]{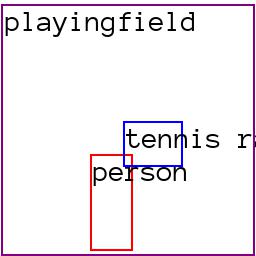}}
& \fbox{\includegraphics[width=1in]{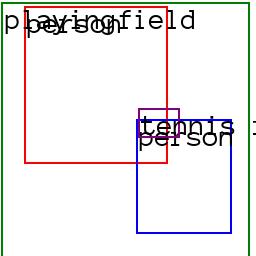}}
& \fbox{\includegraphics[width=1in]{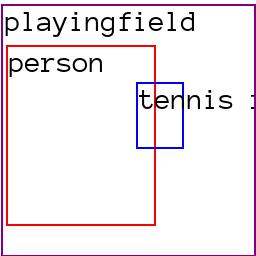}}
\\
\multicolumn{5}{c}{ \{person, playingfield, tennis racket\} }
\vspace{3mm}
\\
\fbox{\includegraphics[width=1in]{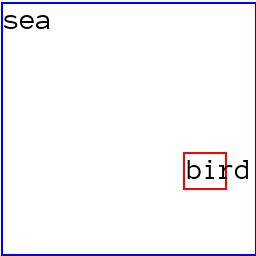}}
&\fbox{\includegraphics[width=1in]{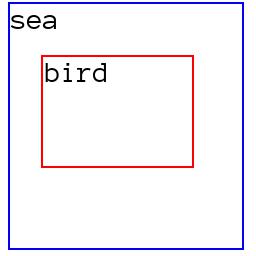}}
& \fbox{\includegraphics[width=1in]{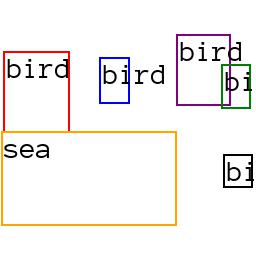}}
& \fbox{\includegraphics[width=1in]{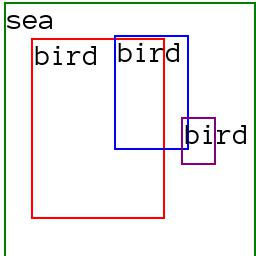}}
& \fbox{\includegraphics[width=1in]{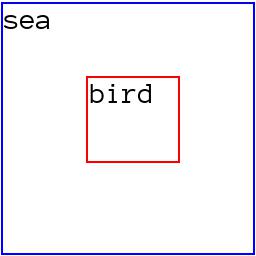}}
\\
\multicolumn{5}{c}{ \{bird, sea\} }
\vspace{3mm}
\\
\fbox{\includegraphics[width=1in]{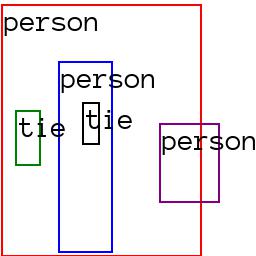}}
&\fbox{\includegraphics[width=1in]{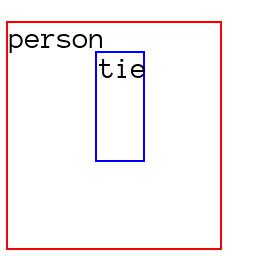}}
& \fbox{\includegraphics[width=1in]{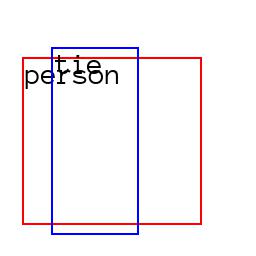}}
& \fbox{\includegraphics[width=1in]{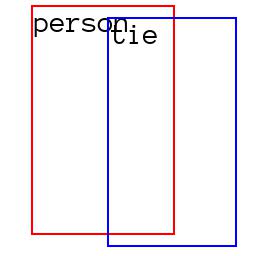}}
& \fbox{\includegraphics[width=1in]{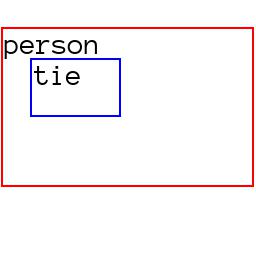}}
\\
\multicolumn{5}{c}{ \{person, tie\} }
\vspace{3mm}
\\
\end{tabular}
\caption{\textbf{Layout generation using LayoutVAE.} We show 5 randomly sampled layouts for each input label set.} 
\label{fig:layout_gen}
\end{figure*}

\autoref{fig:layout_gen} shows examples of diverse layouts generated using LayoutVAE.

\clearpage
\subsection{Examples for Bounding Box Generation}
\label{sec:supp_bbox_generation}

\begin{figure*}[h]
\centering
\pdftooltip{\begin{tabular}{ccccccc}
\hspace{-4mm} \includegraphics[width=1.2in]{figs/app_next_bb/000000000872_1h_person.jpeg}
& \hspace{-6.5mm} \includegraphics[width=1.2in]{figs/app_next_bb/000000000872_2h_person.jpeg}
& \hspace{-6.5mm} \includegraphics[width=1.2in]{figs/app_next_bb/000000000872_3h_sports_ball.jpeg}
& \hspace{-6.5mm} \includegraphics[width=1.2in]{figs/app_next_bb/000000000872_4h_baseball_glove.jpeg}
& \hspace{-6.5mm} \includegraphics[width=1.2in]{figs/app_next_bb/000000000872_5h_playingfield.jpeg}
& \hspace{-6.5mm} \includegraphics[width=1.2in]{figs/app_next_bb/000000000872_6h_tree-merged.jpeg}
\vspace{-2.5mm}
\\
\hspace{-4mm}
\includegraphics[width=1.2in]{figs/app_next_bb/000000000872_1b_person.jpeg}
& \hspace{-6.5mm} \includegraphics[width=1.2in]{figs/app_next_bb/000000000872_2b_person.jpeg}
& \hspace{-6.5mm} \includegraphics[width=1.2in]{figs/app_next_bb/000000000872_3b_sports_ball.jpeg}
& \hspace{-6.5mm} \includegraphics[width=1.2in]{figs/app_next_bb/000000000872_4b_baseball_glove.jpeg}
& \hspace{-6.5mm} \includegraphics[width=1.2in]{figs/app_next_bb/000000000872_5b_playingfield.jpeg}
& \hspace{-6.5mm} \includegraphics[width=1.2in]{figs/app_next_bb/000000000872_6b_tree-merged.jpeg}
\vspace{-1.5mm}
\\
\hspace{-4mm} person & \hspace{-6.5mm} person & \hspace{-6.5mm} sports ball & \hspace{-6.5mm} baseball glove & \hspace{-6.5mm} playingfield & \hspace{-6.5mm} tree-merged
\\
\hspace{-4mm} \includegraphics[width=1.2in]{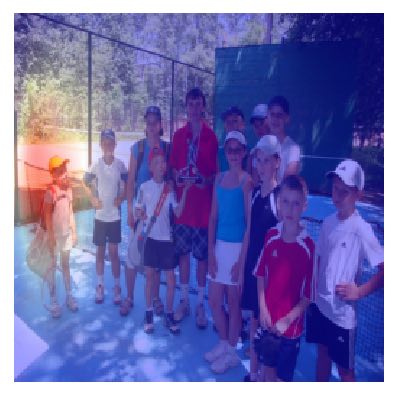}
& \hspace{-6.5mm} \includegraphics[width=1.2in]{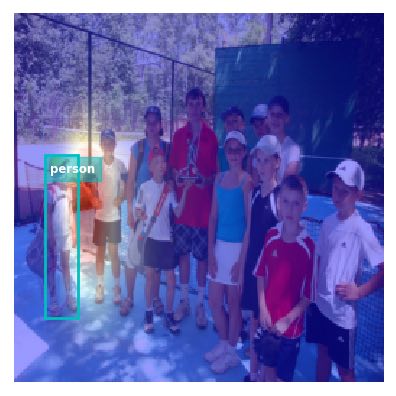}
& \hspace{-6.5mm} \includegraphics[width=1.2in]{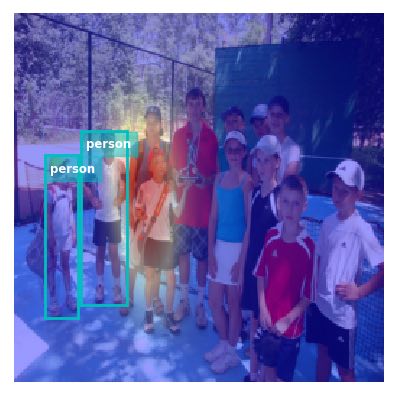}
& \hspace{-6.5mm} \includegraphics[width=1.2in]{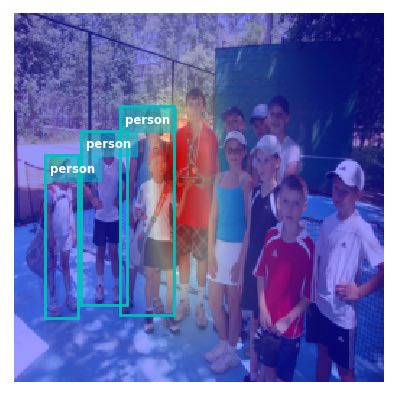}
& \hspace{-6.5mm} \includegraphics[width=1.2in]{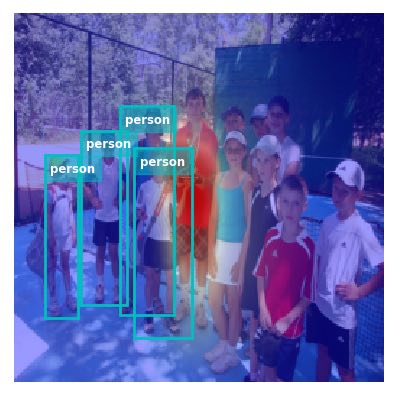}
& \hspace{-6.5mm} \includegraphics[width=1.2in]{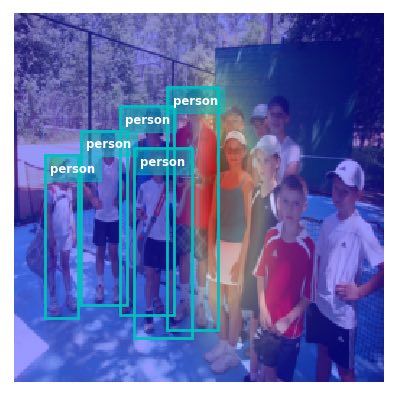}
\vspace{-2.5mm}
\\
\hspace{-4mm} \includegraphics[width=1.2in]{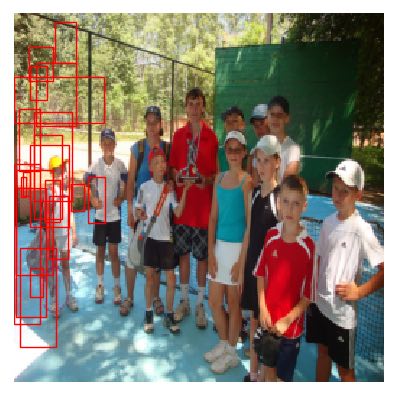}
& \hspace{-6.5mm} \includegraphics[width=1.2in]{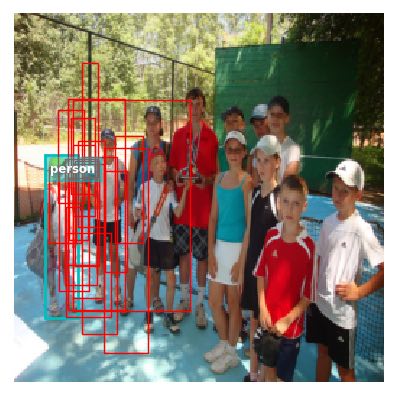}
& \hspace{-6.5mm} \includegraphics[width=1.2in]{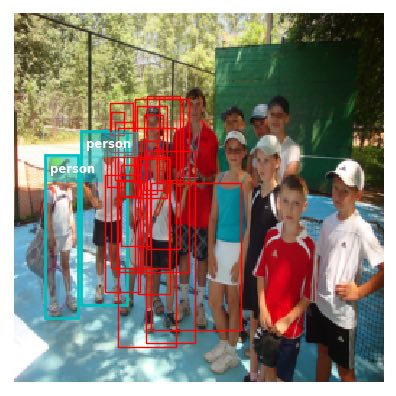}
& \hspace{-6.5mm} \includegraphics[width=1.2in]{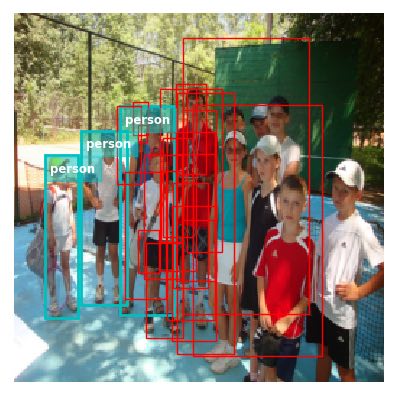}
& \hspace{-6.5mm} \includegraphics[width=1.2in]{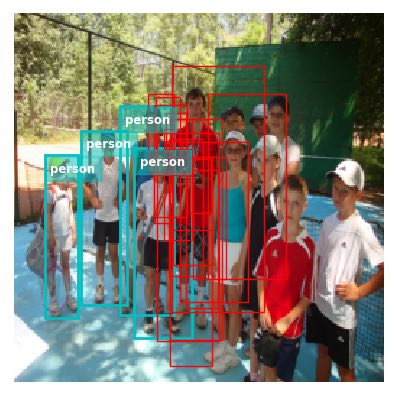}
& \hspace{-6.5mm} \includegraphics[width=1.2in]{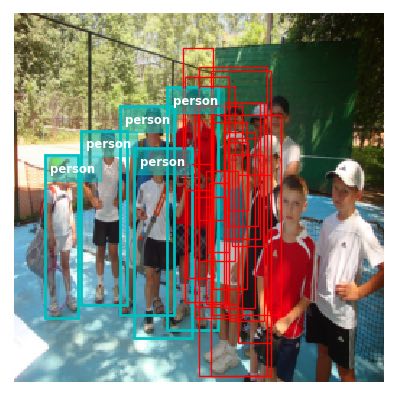}
\vspace{-1.5mm}
\\
\hspace{-4mm} person & \hspace{-6.5mm} person & \hspace{-6.5mm} person & \hspace{-6.5mm} person & \hspace{-6.5mm} person & \hspace{-6.5mm} person 
\end{tabular}}{Images from COCO dataset, originally from Flickr.com. \textCR \textCR First image available at http://farm9.staticflickr.com/8447/7805810128_605424213d_z.jpg, licensed under CC BY 2.0 (https://creativecommons.org/licenses/by/2.0/)
\textCR \textCR Second image available at http://farm5.staticflickr.com/4115/4906536419_6113bd7de4_z.jpg, licensed under CC BY 2.0 (https://creativecommons.org/licenses/by/2.0/)}
\caption{\textbf{Importance of conditioning information.} We present two examples here both of which have \textit{person} as their first label. We see that bounding boxes for the first label  \textit{person} (first column) are close to the center in the first example, whereas they are smaller and close to the left side of the scene in the second example. This is because BBoxVAE gets count of \textit{persons} (2 for the first example, and 12 for the second examples) in its conditioning information, which prompts the model to predict bounding boxes appropriately.
}
\label{fig:next_bb_person}
\end{figure*}

\begin{figure*}[h]
\centering
\pdftooltip{\begin{tabular}{ccccccc}
\hspace{-4mm} \includegraphics[width=1.2in]{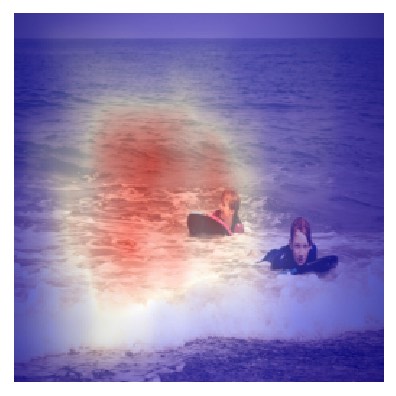}
& \hspace{-6.5mm} \includegraphics[width=1.2in]{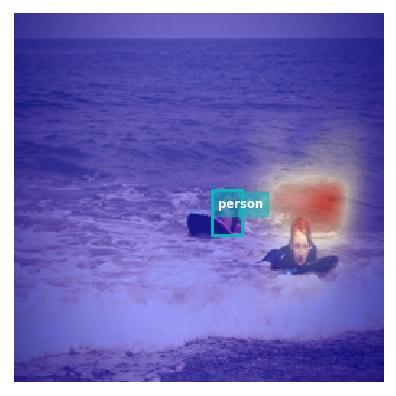}
& \hspace{-6.5mm} \includegraphics[width=1.2in]{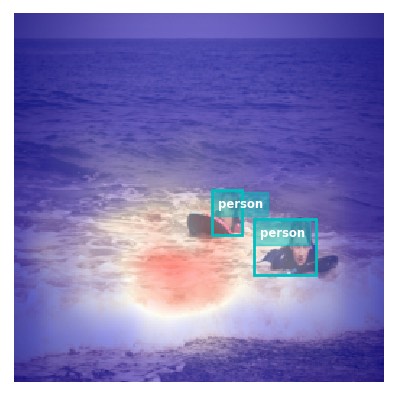}
& \hspace{-6.5mm} \includegraphics[width=1.2in]{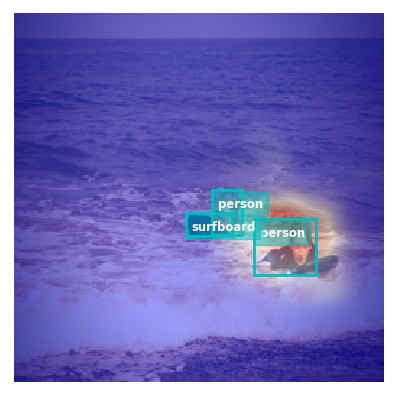}
& \hspace{-6.5mm} \includegraphics[width=1.2in]{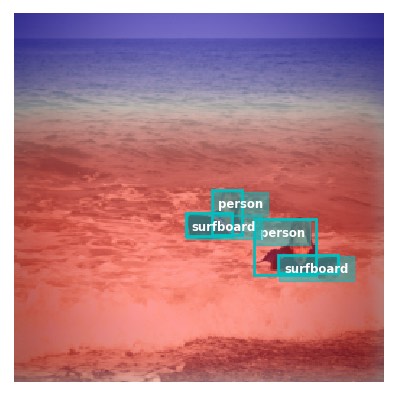}
& \hspace{-6.5mm} \includegraphics[width=1.2in]{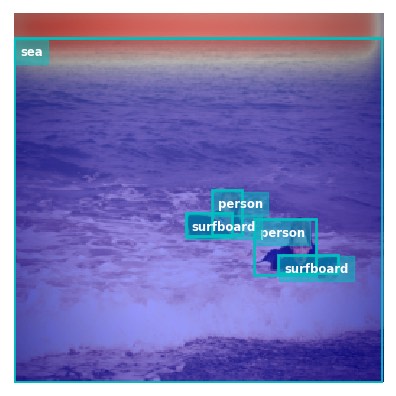}
\vspace{-2.5mm}
\\
\hspace{-4mm} \includegraphics[width=1.2in]{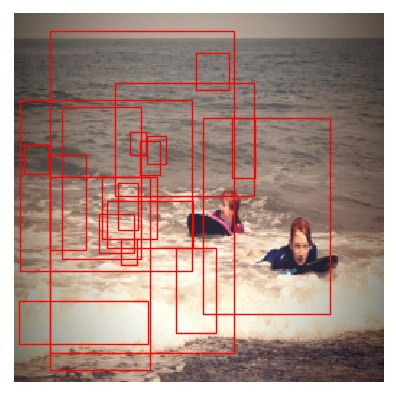}
& \hspace{-6.5mm} \includegraphics[width=1.2in]{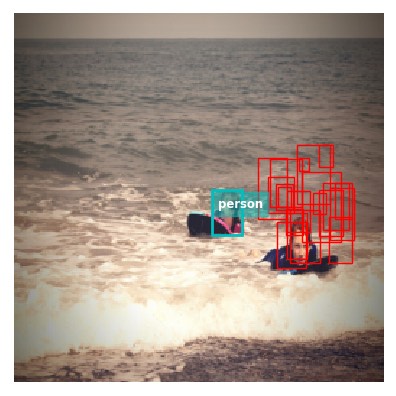}
& \hspace{-6.5mm} \includegraphics[width=1.2in]{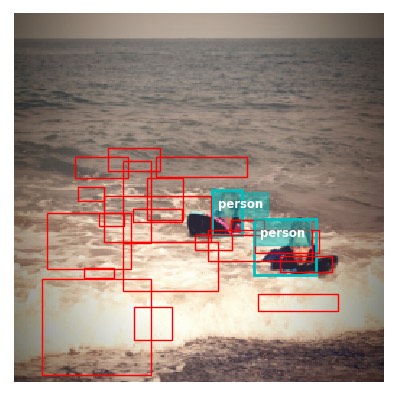}
& \hspace{-6.5mm} \includegraphics[width=1.2in]{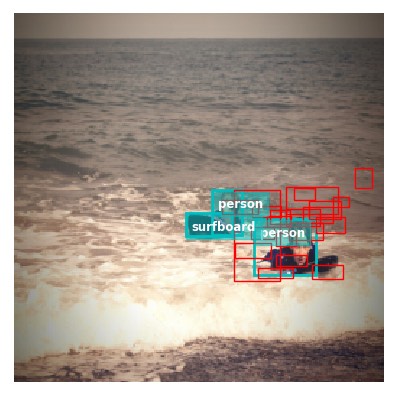}
& \hspace{-6.5mm} \includegraphics[width=1.2in]{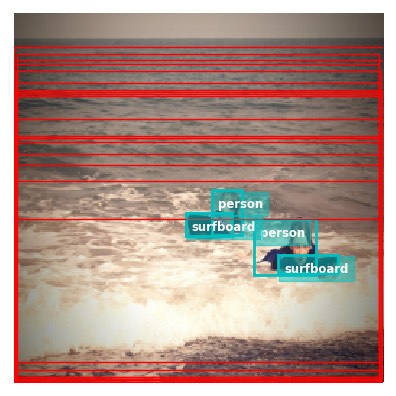}
& \hspace{-6.5mm} \includegraphics[width=1.2in]{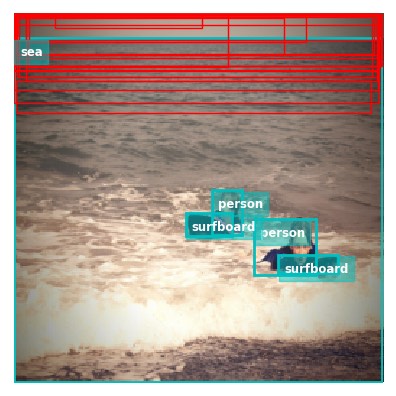}
\vspace{-1.5mm}
\\
\hspace{-4mm} person & \hspace{-6.5mm} person & \hspace{-6.5mm} surfboard & \hspace{-6.5mm} surfboard & \hspace{-6.5mm} sea & \hspace{-6.5mm} sky
\\
\hspace{-4mm} \includegraphics[width=1.2in]{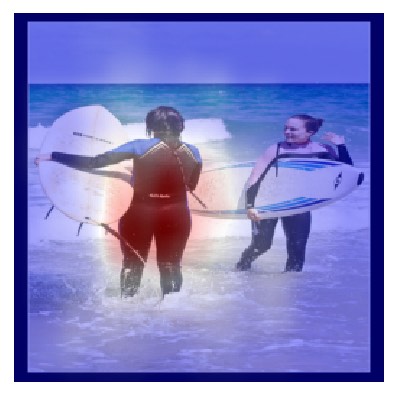}
& \hspace{-6.5mm} \includegraphics[width=1.2in]{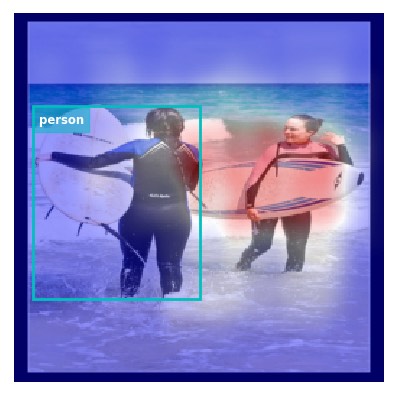}
& \hspace{-6.5mm} \includegraphics[width=1.2in]{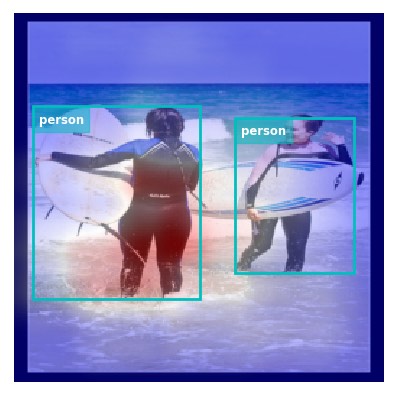}
& \hspace{-6.5mm} \includegraphics[width=1.2in]{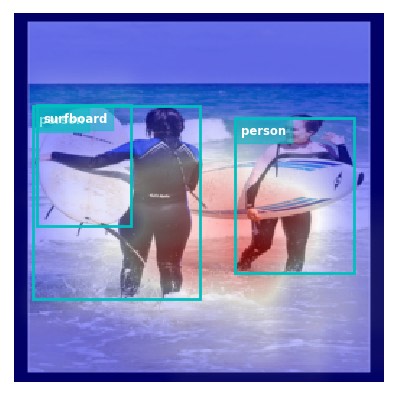}
& \hspace{-6.5mm} \includegraphics[width=1.2in]{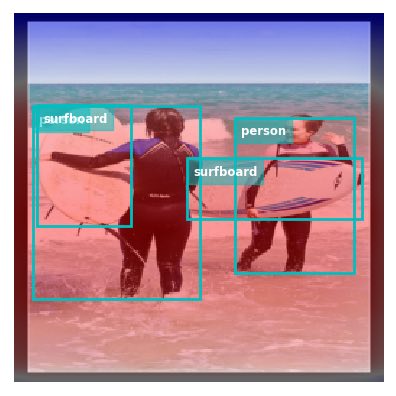}
& \hspace{-6.5mm} \includegraphics[width=1.2in]{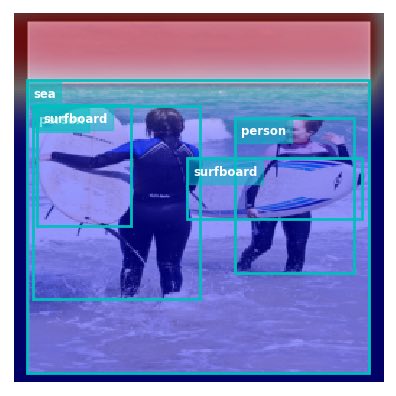}
\vspace{-2.5mm}
\\
\hspace{-4mm} \includegraphics[width=1.2in]{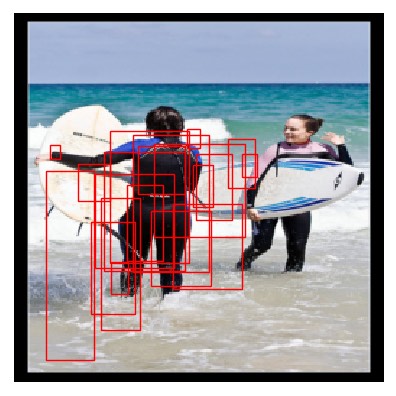}
& \hspace{-6.5mm} \includegraphics[width=1.2in]{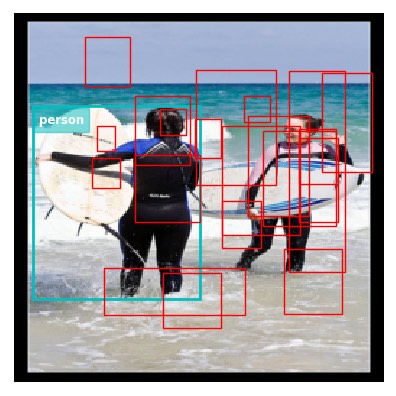}
& \hspace{-6.5mm} \includegraphics[width=1.2in]{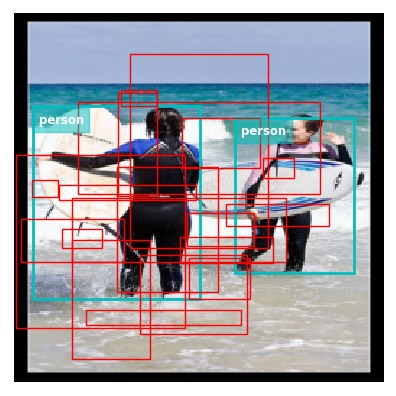}
& \hspace{-6.5mm} \includegraphics[width=1.2in]{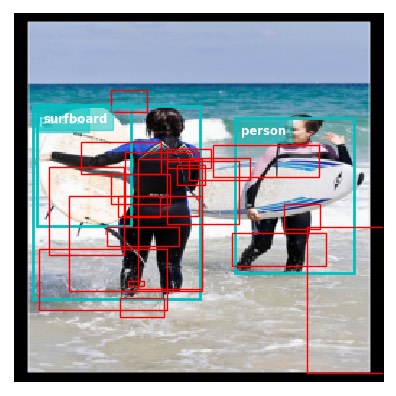}
& \hspace{-6.5mm} \includegraphics[width=1.2in]{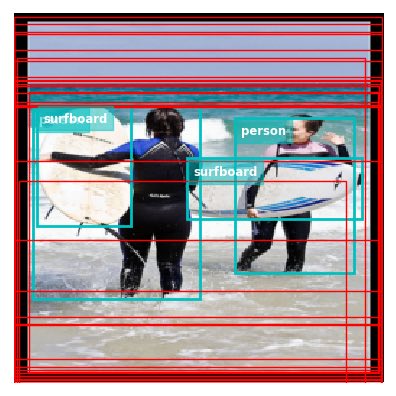}
& \hspace{-6.5mm} \includegraphics[width=1.2in]{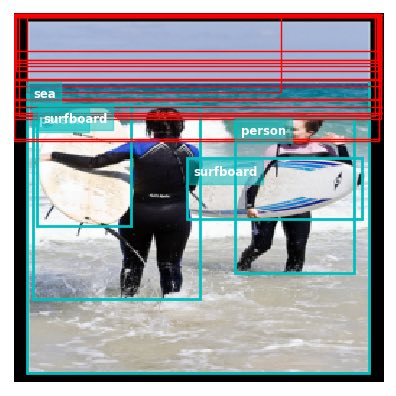}
\vspace{-1.5mm}
\\
\hspace{-4mm} person & \hspace{-6.5mm} person & \hspace{-6.5mm} surfboard & \hspace{-6.5mm} surfboard & \hspace{-6.5mm} sea & \hspace{-6.5mm} sky 
\\
\hspace{-4mm} \includegraphics[width=1.2in]{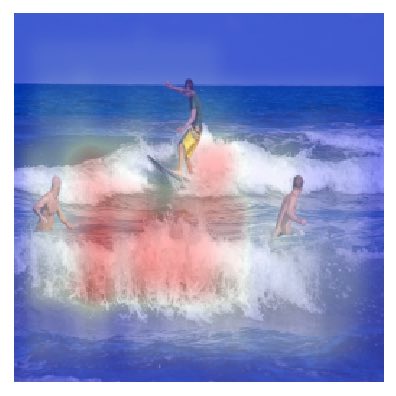}
& \hspace{-6.5mm} \includegraphics[width=1.2in]{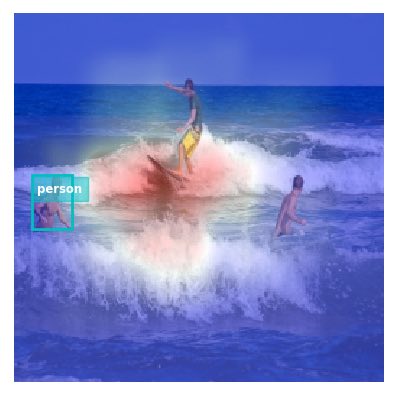}
& \hspace{-6.5mm} \includegraphics[width=1.2in]{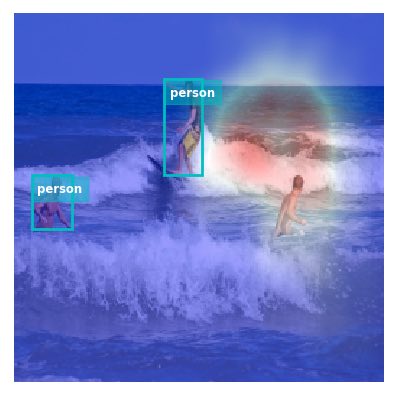}
& \hspace{-6.5mm} \includegraphics[width=1.2in]{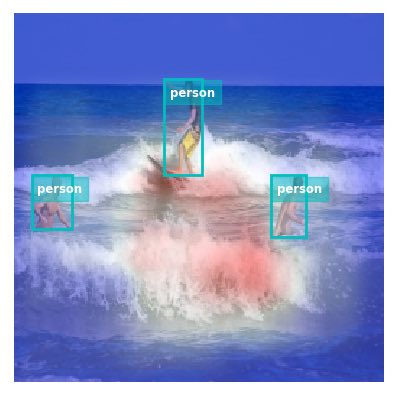}
& \hspace{-6.5mm} \includegraphics[width=1.2in]{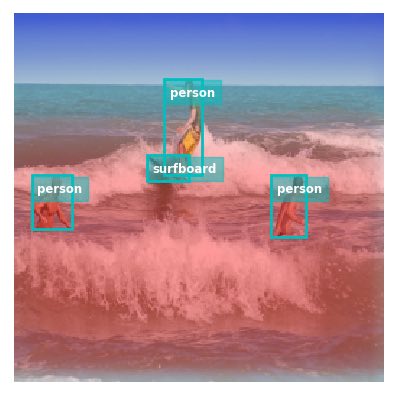}
& \hspace{-6.5mm} \includegraphics[width=1.2in]{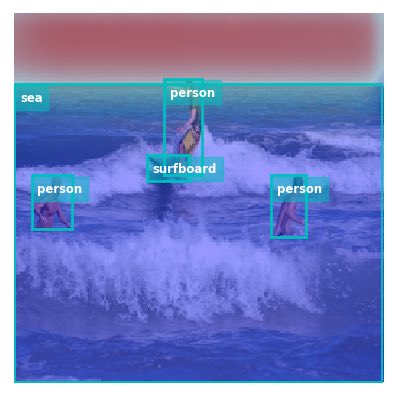}
\vspace{-2.5mm}
\\
\hspace{-4mm} \includegraphics[width=1.2in]{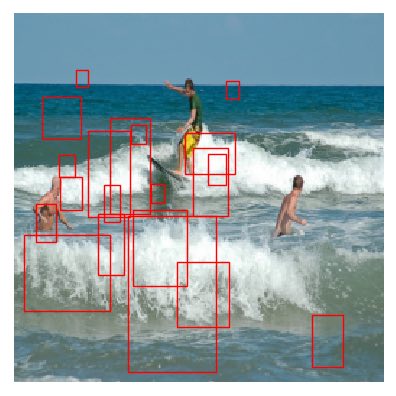}
& \hspace{-6.5mm} \includegraphics[width=1.2in]{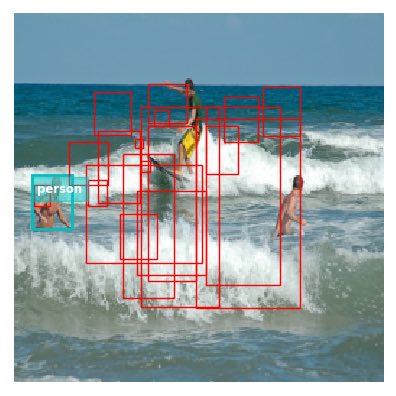}
& \hspace{-6.5mm} \includegraphics[width=1.2in]{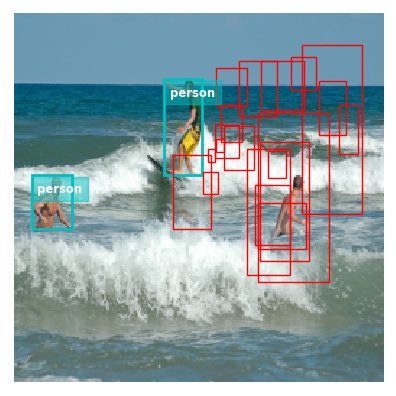}
& \hspace{-6.5mm} \includegraphics[width=1.2in]{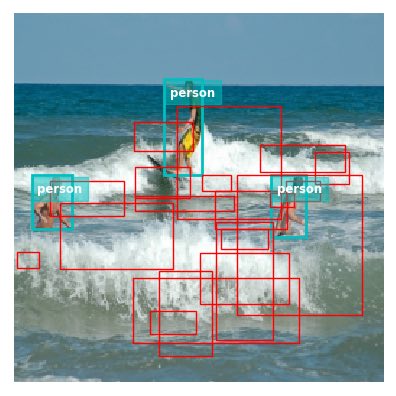}
& \hspace{-6.5mm} \includegraphics[width=1.2in]{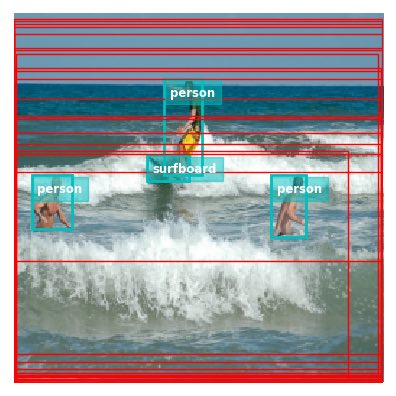}
& \hspace{-6.5mm} \includegraphics[width=1.2in]{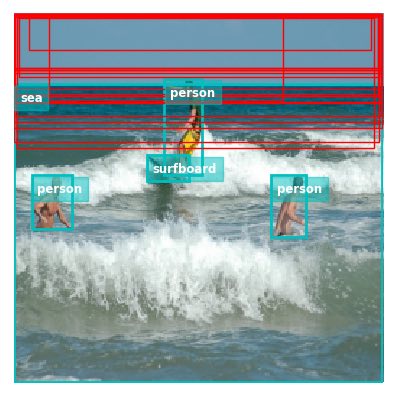}
\vspace{-1.5mm}
\\
\hspace{-4mm} person & \hspace{-6.5mm} person & \hspace{-6.5mm} person & \hspace{-6.5mm} surfboard & \hspace{-6.5mm} sea & \hspace{-6.5mm} sky 
\end{tabular}}
{Images from COCO dataset, originally from Flickr.com. 
\textCR \textCR First image available at http://farm5.staticflickr.com/4094/4891067097_caa3cccb18_z.jpg, licensed under CC BY 2.0 (https://creativecommons.org/licenses/by/2.0/)
\textCR \textCR Second image available at http://farm3.staticflickr.com/2122/2508706790_85dd3ed407_z.jpg, licensed under CC BY-NC-SA 2.0 (https://creativecommons.org/licenses/by-nc-sa/2.0/)
\textCR \textCR Third image available at http://farm1.staticflickr.com/117/310847809_64c76b94c2_z.jpg, licensed under CC BY-NC-ND 2.0 (https://creativecommons.org/licenses/by-nc-nd/2.0/)}

\caption{\textbf{Test examples with labels person, surfboard, sea.} We show steps of bounding box generation for test set samples in the same manner as in Figure 7 from the main paper.} 
\label{fig:next_bb_surfboard2}
\end{figure*}

\begin{figure*}
\centering
\pdftooltip{\begin{tabular}{ccccccc}
\hspace{-4mm} \includegraphics[width=1.2in]{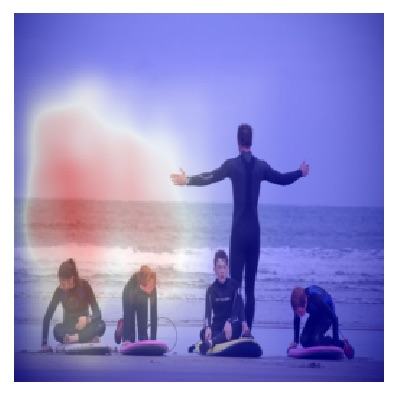} 
& \hspace{-6.5mm} \includegraphics[width=1.2in]{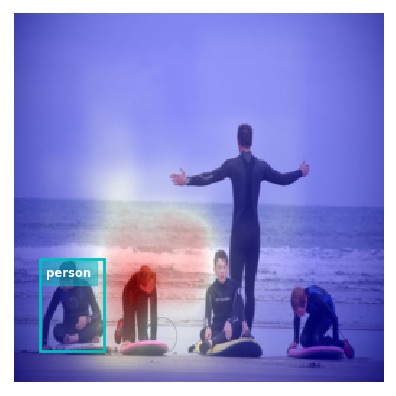} 
& \hspace{-6.5mm} \includegraphics[width=1.2in]{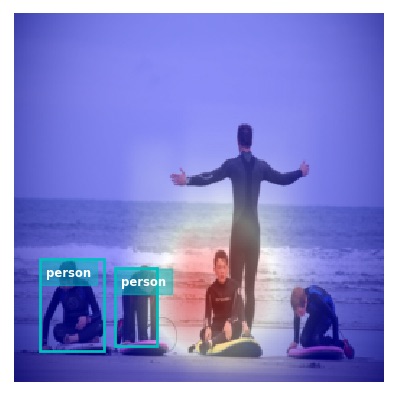} 
& \hspace{-6.5mm} \includegraphics[width=1.2in]{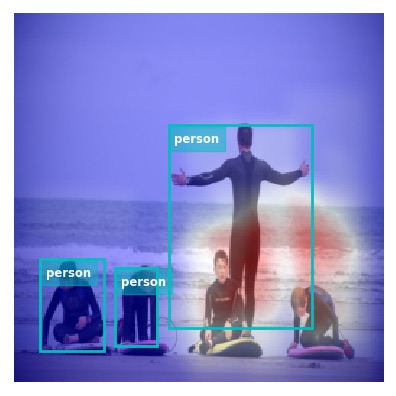} 
& \hspace{-6.5mm} \includegraphics[width=1.2in]{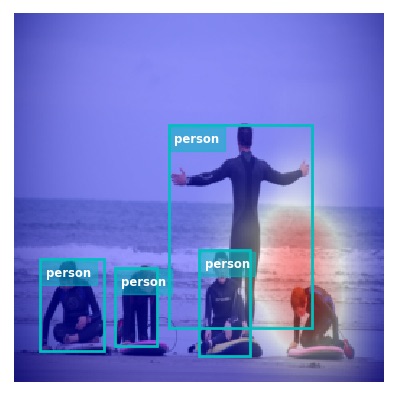} 
& \hspace{-6.5mm} \includegraphics[width=1.2in]{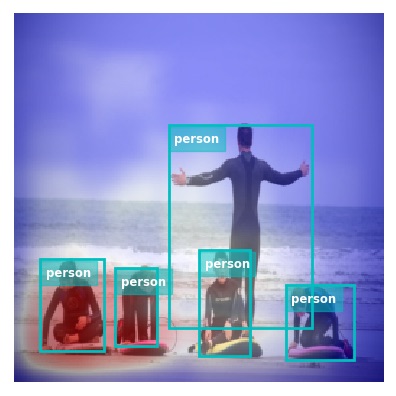} 
\vspace{-2.5mm}
\\
\hspace{-4mm}
\includegraphics[width=1.2in]{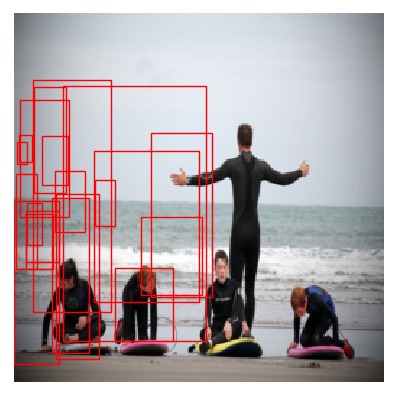} 
& \hspace{-6.5mm} \includegraphics[width=1.2in]{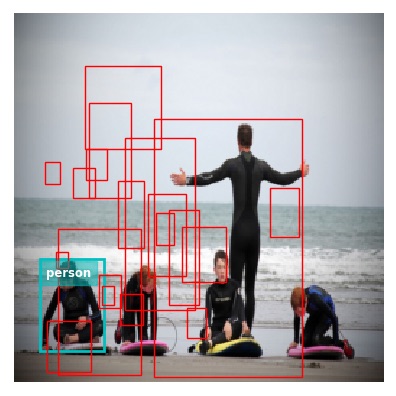} 
& \hspace{-6.5mm} \includegraphics[width=1.2in]{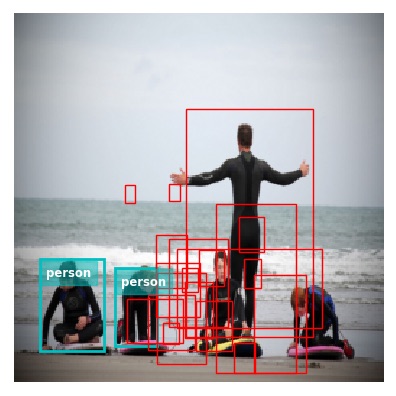} 
& \hspace{-6.5mm} \includegraphics[width=1.2in]{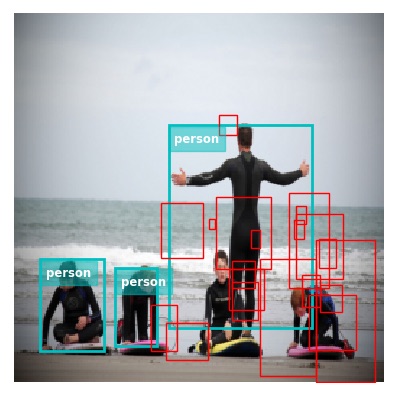} 
& \hspace{-6.5mm} \includegraphics[width=1.2in]{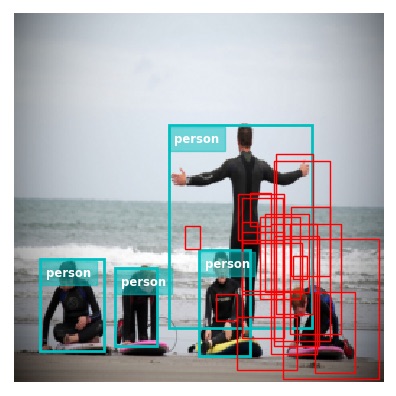} 
& \hspace{-6.5mm} \includegraphics[width=1.2in]{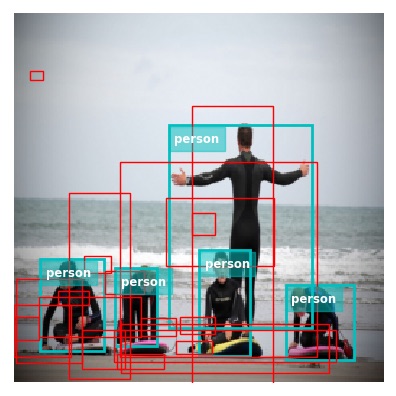} 
\vspace{-1.5mm}
\\
\hspace{-4mm} person & \hspace{-6.5mm} person & \hspace{-6.5mm} person & \hspace{-6.5mm} person & \hspace{-6.5mm} person & \hspace{-6.5mm} surfboard 
\\
\hspace{-4mm} \includegraphics[width=1.2in]{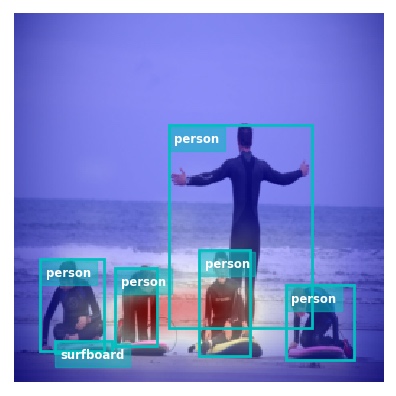} 
& \hspace{-6.5mm} \includegraphics[width=1.2in]{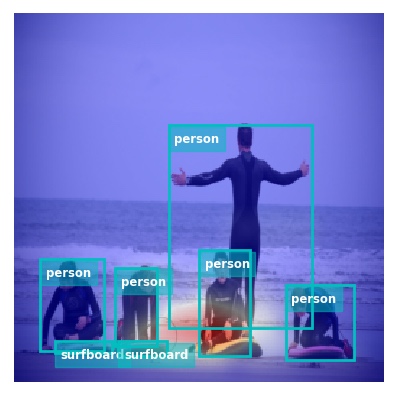}
& \hspace{-6.5mm} \includegraphics[width=1.2in]{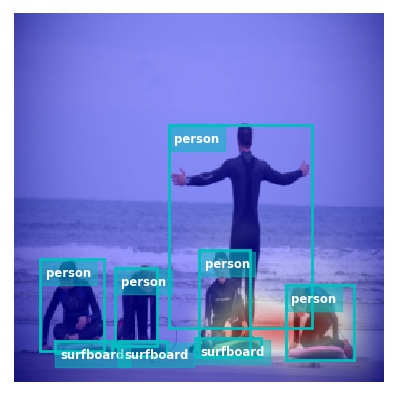} 
& \hspace{-6.5mm} \includegraphics[width=1.2in]{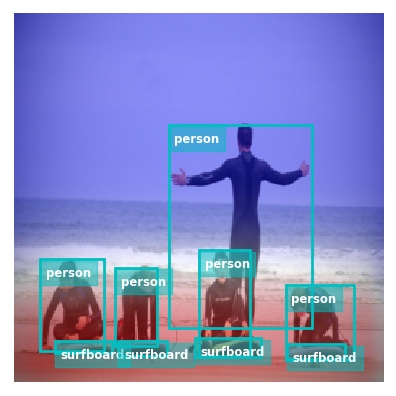} 
& \hspace{-6.5mm} \includegraphics[width=1.2in]{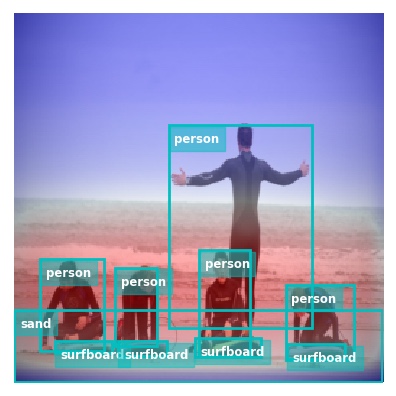} 
& \hspace{-6.5mm} \includegraphics[width=1.2in]{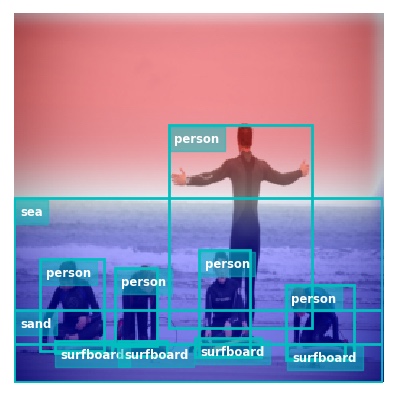} 
\vspace{-2.5mm}
\\
\hspace{-4mm} \includegraphics[width=1.2in]{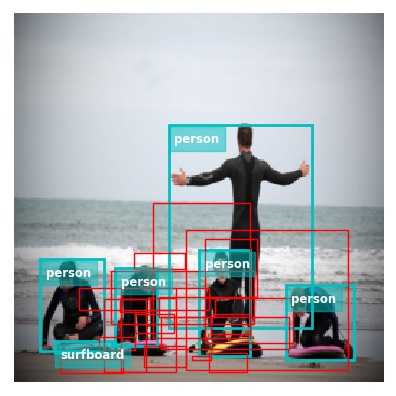} 
& \hspace{-6.5mm} \includegraphics[width=1.2in]{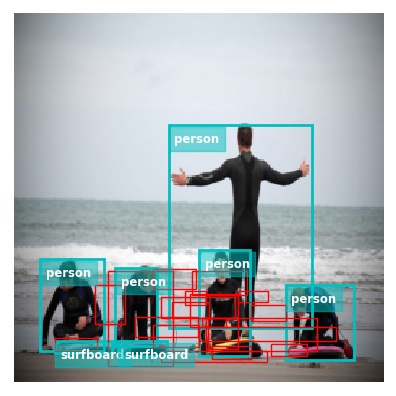} 
& \hspace{-6.5mm} \includegraphics[width=1.2in]{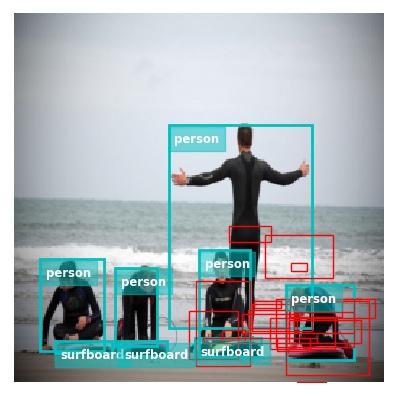} 
& \hspace{-6.5mm} \includegraphics[width=1.2in]{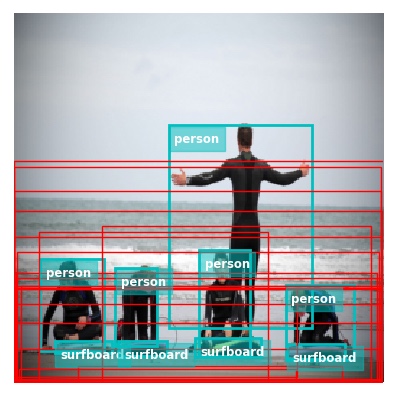} 
& \hspace{-6.5mm} \includegraphics[width=1.2in]{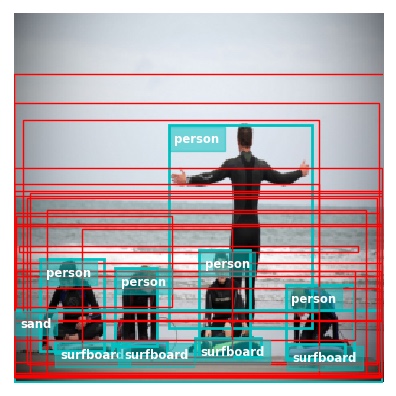} 
& \hspace{-6.5mm} \includegraphics[width=1.2in]{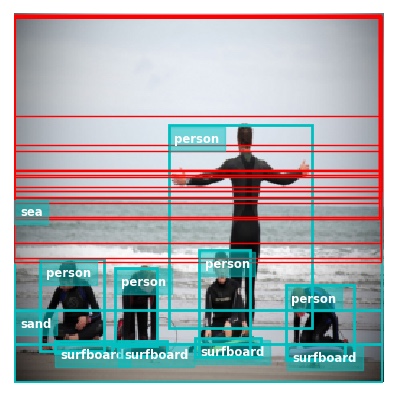} 
\vspace{-1.5mm}
\\
\hspace{-4mm} surfboard & \hspace{-6.5mm} surfboard & \hspace{-6.5mm} surfboard & \hspace{-6.5mm} sand & \hspace{-6.5mm} sea & \hspace{-6.5mm} sky 
\end{tabular}}
{Image from COCO dataset, originally from Flickr.com and available at http://farm7.staticflickr.com/6080/6038911779_36e4613d84_z.jpg, licensed under CC BY 2.0 (https://creativecommons.org/licenses/by/2.0/)}

\vspace{-3mm}
\caption{\textbf{An example with more objects in the scene.} We show steps of bounding box generation for test set samples in the same manner as in Figure 7 from the main paper.
}
\label{fig:next_bb_surfboard3}
\end{figure*}

\autoref{fig:next_bb_person} presents examples that showcase the ability of LayoutVAE to use conditioning information to predict plausible bounding boxes.
We present additional examples of stochastic bounding box generation for test samples that have labels \textit{person}, \textit{surfboard} and \textit{sea} in \autoref{fig:next_bb_surfboard2} and \autoref{fig:next_bb_surfboard3}.

\end{document}